\documentclass[11pt]{article}

\usepackage[final]{acl}

\usepackage{times}
\usepackage{latexsym}
\usepackage[T1]{fontenc}
\usepackage[utf8]{inputenc}
\usepackage{microtype}
\usepackage{inconsolata}

\usepackage{amsmath}
\usepackage{amssymb}
\usepackage{graphicx}
\usepackage{placeins}
\usepackage{booktabs}
\usepackage{multirow}
\usepackage{xcolor}
\usepackage{subcaption}
\usepackage{algorithm}
\usepackage{algorithmic}
\usepackage{xspace}
\usepackage{enumitem}
\usepackage{pifont}
\usepackage{fontawesome5}
\usepackage{twemojis}
\usepackage{tcolorbox}
\usepackage{cleveref}

\tcbuselibrary{breakable}
\newtcolorbox{promptbox}[1][]{
  colback=gray!5,
  colframe=black!75,
  fonttitle=\bfseries\small,
  boxrule=0.5pt,
  arc=2pt,
  left=6pt,
  right=6pt,
  top=4pt,
  bottom=4pt,
  breakable,
  #1
}

\newcommand{\ours}{HT-Coreset\xspace}
\newcommand{\naive}{Na\"ive Coreset\xspace}
\newcommand{\bench}{AMLworld-Compact\xspace}
\newcommand{\hiSmall}{HI-Small\xspace}
\newcommand{\liSmall}{LI-Small\xspace}

\newcommand{\fs}{ICL-FS\xspace}
\newcommand{\zs}{ICL-ZS\xspace}
\newcommand{\verif}{ICL-V\xspace}

\title{AMLworld-Compact: Importance-Weighted Downsampling \\
for Cost-Effective LLM Evaluation and Error Diagnosis}

\author{
  \parbox{\linewidth}{\centering
    \textbf{Natapong~Nitarach}\textsuperscript{1},
    \textbf{Phume~Ngampornsukswadi}\textsuperscript{2},
    \textbf{Pittawat~Taveekitworachai}\textsuperscript{1}\\
    \textbf{Surapon~Nonesung}\textsuperscript{1},
    \textbf{Warit~Sirichotedumrong}\textsuperscript{1}\\
    \textbf{Kunat~Pipatanakul}\textsuperscript{1}\\[4pt]
    \normalfont
    \textsuperscript{1}SCB DataX, SCBX Group \quad
    \textsuperscript{2}Independent Researcher}
}

\begin{document}
\maketitle

\begin{abstract}
Evaluating large language models (LLMs) on AMLworld's smallest test splits
is costly because each of over a million targets requires a separate
graph prompt. Sampling reduces this cost, but small balanced subsets
change unweighted precision and F1.
We introduce \bench{}, an evaluation set that retains every illicit
test transaction and samples benign transactions from supervised-score
strata. Its HT-Coreset protocol reduces evaluation targets by $271\times$
and $611\times$ on \hiSmall and \liSmall.
Horvitz--Thompson weights give unbiased full-test confusion-count
estimates for fixed predictions; recall is exact, while precision and
F1 are ratio estimates.
In a held-out study using Graph Feature Preprocessor (GFP) features,
one boosted-tree family defines strata for evaluating the other over
$500$ draws per configuration.
At the released sizes, F1 RMSE is $0.00$--$2.58$ percentage points versus
$4.44$--$7.41$ for uniform sampling with HT weights; conservative
intervals remain broad.
Under the evaluated graph prompts, seven open-weight LLMs have weighted
precision near the predict-all-illicit baseline.
In an outcome-stratified audit of $1{,}000$ \hiSmall few-shot outputs
from one run per model, lexical checks pass for $84.5\%$ of correct and
$56.2\%$ of incorrect binary verdicts, so these markers do not establish
correctness.
We release the evaluation sets, construction code, and audit tools.%
%
\footnote{\raggedright
\twemoji{1f917}\,\href{https://huggingface.co/datasets/typhoon-ai/AMLworldCompactEval}{\nolinkurl{typhoon-ai/AMLworldCompactEval}}\\
\faGithub\,\href{https://github.com/nat-nischw/AMLworld-Compact}{\nolinkurl{github.com/nat-nischw/AMLworld-Compact}}}%
%
\end{abstract}

\section{Introduction}
\label{sec:intro}

\begin{figure}[t]
\centering
\includegraphics[width=\columnwidth]{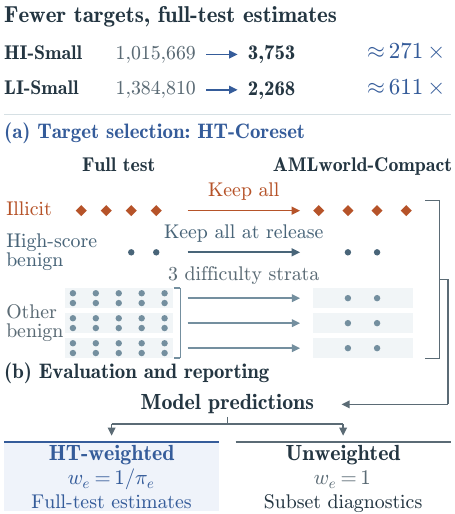}
\caption{HT-Coreset (a) retains illicit targets (diamonds) and samples
benign targets (circles). High-score benign targets form one stratum,
defined by the construction ensemble; all fit at the released budgets.
The other benign targets form three difficulty terciles (shaded rows).
In (b), predictions on the targets selected in (a) are weighted by $1/\pi_e$ for full-test
estimates or counted equally for subset diagnostics; $\pi_e$ is the
inclusion probability. Markers represent transaction targets;
marker counts are illustrative.}
\label{fig:evaluation_overview}
\end{figure}

Anti-money laundering (AML) detection asks whether a transaction
belongs to illicit activity. The evidence can span several accounts
and transfers: a gather-scatter pattern, for example, collects funds
at one account before redistributing them. A large language model
(LLM) can receive this neighborhood as graph text, but evaluating
each transaction separately requires one prompt per target.
The test partitions of AMLworld's two smallest
splits~\citep{altman2023realistic} each contain over a million
transactions, with illicit rates of $0.12\%$ on \hiSmall and
$0.05\%$ on \liSmall. A single pass on either split requires over
a million graph prompts; comparing models and prompt variants
multiplies this workload.

A small class-proportional sample contains few illicit cases,
limiting recall and typology analysis. A balanced sample increases
their share but changes unweighted precision and F1. Prior exploratory
work~\citep{pirmorad2025iclaml} evaluates LLMs on a small balanced
AMLworld subset; those scores reflect its enriched illicit proportion.
The challenge is to reduce model calls while keeping rare cases
available for diagnosis and estimating detection at full-test prevalence.

We introduce \textbf{\bench{}}, constructed with \textit{HT-Coreset}.
The protocol retains every illicit test edge, uses supervised scores
to identify difficult benign cases, and samples the remaining benign
edges within difficulty strata. Known inclusion probabilities give
Horvitz--Thompson (HT) weights~\citep{horvitz1952generalization} for
estimating full-test confusion counts. The weights account for
unsampled benign cases; unweighted counts describe errors on the
selected cases. These two summaries use the same selected targets and
predictions (Figure~\ref{fig:evaluation_overview}); precision and F1
are ratio estimates, with possible finite-sample bias.

The released sets contain $3{,}753$ \hiSmall and $2{,}268$ \liSmall
targets, reducing target counts by $271\times$ and $611\times$.
Each target has GFP features~\citep{blanusa2024gfp} and a typed-subgraph
prompt extracted from the full graph: selecting fewer targets does
not reduce the neighborhood available to each selected case.
Our contributions are:
\begin{itemize}[leftmargin=*,itemsep=2pt,topsep=2pt]
\item \textbf{An evaluation set with a full-test reference.}
      The released targets, inclusion weights, features, and graph
      strings let subsequent models use the same cases and estimate
      metrics for the same population. We compare $12$ budgets and
      $7$ alternative samplers, then test sampling error and interval
      coverage with the evaluated booster family held out of
      construction, over $500$ draws per configuration
      (\Cref{sec:method}).
\item \textbf{LLM evaluation and error analysis.}
      Seven open-weight LLMs with the original prompts have weighted
      precision near full-test illicit prevalence. We analyze
      ML--LLM error overlap and find that lexical checks pass for over
      half of the incorrect binary verdicts in the audit sample.
      A supplementary control shows that the label-derived route
      accounts for most of the apparent hybrid gain
      (\Cref{sec:llm,sec:llm:why}).
\end{itemize}

\section{Related Work}\label{sec:related}

\paragraph{LLM evaluation on AML graphs.}
\citet{pirmorad2025iclaml} explores in-context learning on a small
balanced AMLworld subset. We reuse their typed-graph prompt template
with our own neighborhoods and demonstrations. Our evaluation set
adds a different population reference: inclusion weights estimate
full-test detection counts despite the subset's enriched illicit
proportion. We study sampling error and use the selected cases to
analyze detection and written outputs.

\paragraph{Graph inputs and graph retrieval.}
\citet{chen2024llmgraph} study LLMs as feature enhancers and direct
predictors for node classification; \citet{fatemi2024talk} examine
how text encodings affect graph reasoning. Our task is transaction
classification from a fixed $k$-hop graph presented as text.
GraphRAG~\citep{edge2024graphrag} instead retrieves information over
a graph index. Retrieval quality and reasoning over a supplied
neighborhood are distinct evaluation problems.

\paragraph{Efficient evaluation and coreset selection.}
tinyBenchmarks~\citep{polo2024tinybenchmarks} and
metabench~\citep{kipnis2025metabench} include latent-ability methods
that use prior model responses to estimate full-benchmark scores
from small test subsets. Here, known transaction labels identify a rare class
that we retain in full; supervised scores allocate the benign
sampling budget. CRAIG~\citep{mirzasoleiman2020coresets} and
CCS~\citep{zheng2023coveragecentric} select training coresets,
whereas our subset is used for evaluation.
\citet{fogliato2024reducedlabel} study stratified sampling with
design-based and model-assisted estimators for classification
evaluation. Our Neyman-allocation baseline~\citep{neyman1934allocation}
adapts their framework to binary classification, sampling both classes;
HT-Coreset retains all illicit edges and
weights sampled benign edges. The comparisons in the next section
test these choices at matched budgets.

\section{Constructing \bench{}}
\label{sec:method}

HT-Coreset takes the full test partition and supervised scores as
inputs. It selects transaction targets and records their inclusion
probabilities for subsequent evaluation.

\subsection{Setting}
\label{sec:method:setup}

\paragraph{Evaluation population.}
We use \hiSmall and \liSmall, the two smallest splits of
AMLworld~\citep{altman2023realistic}. Its labeled synthetic
transaction graphs let us compare sampled estimates with known
full-test metrics. Each edge is labeled \textit{illicit} or
\textit{benign}; illicit edges may carry one of $8$ typology
annotations (\Cref{app:dataset}). We first partition the source CSV
rows $60/20/20$ for training, validation, and testing in file order.
HT-Coreset then selects targets only from the test partition.
Timestamps overlap across partitions, so this split is not strictly
chronological. The released sets use one illicit edge to two benign
edges ($1{:}2$), the smallest budget meeting our tested construction
criterion.

\paragraph{Construction scores.}
Scores guide the benign budget toward likely false positives, which
can strongly affect precision at low illicit prevalence.
We use LightGBM~\citep{ke2017lightgbm}, XGBoost~\citep{chen2016xgboost},
and GCPAL~\citep{lu2024gcpal} with Graph Feature Preprocessor
(GFP) features~\citep{blanusa2024gfp} to define the strata.
The boosters train on the training partition. The archived GCPAL
checkpoints use a random fine-tuning split that overlaps the test
set. We retain the frozen construction design and its inclusion
weights; task comparisons use only the two boosters.
\Cref{app:non_llm} gives feature dimensions, architectures, training
settings, and split provenance.

\subsection{Class-Proportional Downsampling}
\label{sec:method:naive}

\naive preserves the class ratio while stratifying illicit edges
by typology and difficulty. At small budgets, few illicit edges
remain for recall estimation. The smallest feasible draw found for
\hiSmall in our constrained search contains $641$K edges
(absolute F1 deviation at most $5$ pp; \Cref{app:naive_vs_ht}). Retaining all
illicit cases instead allows a much smaller benign budget, with
importance weights accounting for unequal benign-edge selection probabilities.

\subsection{\ours}
\label{sec:method:coreset}

\paragraph{Stage 1: retain all illicit cases.}
Keeping every illicit edge preserves recall information and all
$8$ typologies rather than relying on a small minority sample.

\paragraph{Stage 2: allocate the benign budget.}
To retain likely false positives, we score each edge with
$\bar p(e)$, the mean across $3$ scorers and $5$ seeds.
We define the difficulty score as
\[
  d(e) = 1 - 2\,\lvert \bar p(e) - 0.5 \rvert .
\]
Difficulty is highest at a mean probability of $0.5$.
Hard negatives are benign edges with mean probabilities of at least
$0.40$ on \hiSmall or $0.24$ on \liSmall. These cutoffs are half
the ensemble's respective operating thresholds.

Up to $30\%$ of the benign budget is reserved for this high-score
stratum. We use the remaining budget to sample other benign edges
from three difficulty strata (terciles). These four benign strata give every
benign edge a positive inclusion probability. At the released
budgets, all hard negatives fit, with $363$ on \hiSmall
($14.5\%$ of its benign sample) and $269$ on \liSmall ($17.8\%$).

We select the smallest of $12$ benign budgets whose mean absolute
HT-weighted deviations in precision ($P$), recall ($R$), and F1
are each at most $5$ percentage points (pp) over $50$ draws.
The selected budget is twice the illicit count, giving the $1{:}2$
ratio used for LLM evaluation. This is the maximum reduction among
the budgets tested.

\paragraph{Stage 3: estimate full-test counts.}
\label{sec:method:estimation}
Each selected edge receives weight $w_e=1/\pi_e$, the inverse of
its inclusion probability. Retained illicit edges have weight one.
Summing weights over TP, FP, and FN gives estimated full-test
confusion counts. Precision, recall, and F1 are computed from these
counts. For fixed predictions and known positive inclusion
probabilities, the counts are unbiased. Complete illicit retention
makes recall exact; precision and F1 are ratio estimates with
possible finite-sample bias and sampling variance
(\Cref{app:estimation}). \Cref{alg:ht_coreset} gives the procedure.

\paragraph{Two summaries of the same predictions.}
Both reports use the same targets selected by HT-Coreset and the same
predictions. HT-weighted scoring estimates full-test performance;
unweighted scoring sets $w_e=1$ to describe the enriched subset.
A predictor that flags every edge has $33.3\%$ precision and
$50.0\%$ F1 on the released $1{:}2$ subset, but $0.123\%$
precision and $0.246\%$ F1 on the full \hiSmall test split.
HT weighting recovers the full-test predict-all reference.

\subsection{Typed-Graph Serialization for LLMs}
\label{sec:method:serial}

Supervised models consume a GFP feature row per target. Its pattern
histograms mirror five of the eight typologies, giving explicit
pattern counts to the supervised models. LLMs instead receive the typed-graph text format of
\citet{pirmorad2025iclaml}. Starting at the target's source account,
our extractor runs breadth-first search over the full directed
transaction graph to depth $k{=}2$, considering incoming and
outgoing neighbors. It selects at most $50$ unvisited neighboring
accounts per expanded account per hop
(\Cref{app:subgraph,tab:subgraph_params}). Accounts and banks are
typed nodes; transfers and bank memberships are typed relations.
Transaction attributes include amount, currency, payment method,
and timestamp. Few-shot explanations identify the illustrated typologies.

The evaluated prompts retain a literal \texttt{<ID>} placeholder
and do not explicitly identify the target transaction. The target
is absent from $324$ of $3{,}753$ \hiSmall and $297$ of $2{,}268$
\liSmall contexts (\Cref{app:prompt_provenance}). A separately
released target-marked version preserves case IDs and weights,
identifies and includes each target, and corrects the time summary.
It has not been evaluated with an LLM; source-record and
serialization checks appear in \Cref{app:target_integrity}.

The released original graph strings total $272$M characters for
\hiSmall and $168$M for \liSmall, or roughly $68$M and $42$M
graph-text tokens per pass using four characters per token.
Instructions, demonstrations, and outputs are excluded;
actual input counts depend on the tokenizer.

\subsection{Released Evaluation Set}
\label{sec:method:resource}

The release aligns case IDs, binary labels, available typology
annotations, GFP feature rows, graph text, and HT weights. It retains
all $1{,}251$ \hiSmall and $756$ \liSmall illicit edges and all
$8$ typologies (Table~\ref{tab:resource_summary}). Target selection
uses seed $0$ and is reproducible from the original construction
scores.

\begin{table}[t]
\centering
\small
\setlength{\tabcolsep}{4pt}
\begin{tabular}{lrr}
\toprule
 & \textbf{HI-Small} & \textbf{LI-Small} \\
\midrule
Full test edges & $1{,}015{,}669$ & $1{,}384{,}810$ \\
Retained targets & $3{,}753$ & $2{,}268$ \\
Illicit edges (all retained) & $1{,}251$ & $756$ \\
Full-test illicit rate & $0.12\%$ & $0.05\%$ \\
Subset illicit rate & $33.3\%$ & $33.3\%$ \\
Target-count reduction & $271\times$ & $611\times$ \\
Graph-text tokens (est.) & $\sim68$M & $\sim42$M \\
\bottomrule
\end{tabular}
\caption{Full-test and released-set sizes. Each retained edge is
one evaluation target with a neighborhood from the full graph.
Token estimates concern only the released graph strings, using four
characters per token; instructions, demonstrations, and outputs are
excluded. HT weights account for the altered class proportions.}
\label{tab:resource_summary}
\end{table}

\subsection{Metric Agreement and Held-Out Evaluation}
\label{sec:method:ablation}

\begin{figure}[t]
\centering
\includegraphics[width=\columnwidth]{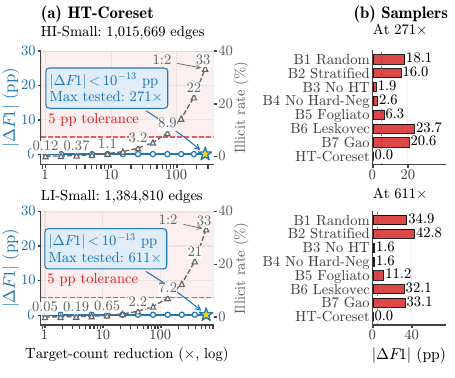}
\caption{Full-split agreement for the construction ensemble.
\textbf{(a)} HT-Coreset has $|\Delta F1|{<}10^{-13}$ pp at
$12$ budgets; stars mark the maximum tested reductions:
$271\times$ on \hiSmall and $611\times$ on \liSmall.
Gray curves and labels show subset illicit percentages
(right axes).
\textbf{(b)} $|\Delta F1|$ at the released budgets ($50$-draw means).
B3/B4 meet the $5$ pp F1 tolerance but exceed it on precision
(Table~\ref{tab:ablation}). Exact agreement follows from retaining
this ensemble's TP/FP/FN contributors at weight one.}
\label{fig:reduction_sweep}
\end{figure}

\paragraph{Construction agreement.}
We compare sampled and full-test edge-level precision, recall, and
F1. Sampling error is $\Delta F1=F1_{\text{subset}}-F1_{\text{full}}$,
and likewise $\Delta P$ and $\Delta R$, in pp.
\Cref{fig:reduction_sweep} shows F1 error across the budget sweep
and seven further samplers. These test ordinary subsampling, benign selection
and weighting, and allocation by model scores, graph topology, or
transaction attributes. We evaluate each over $50$ draws with the
construction ensemble (\Cref{app:construction_comparison}).
Other predictors can flag benign edges outside the retained hard
negatives, so we evaluate them separately.

\paragraph{Holding out a predictor family.}
We construct strata from five-seed LightGBM scores and evaluate
XGBoost, then reverse their roles. Both use GFP features; the
evaluated family does not enter stratum construction and each seed
retains its saved validation-selected threshold.
We keep all illicit edges and compare uniform benign sampling,
difficulty terciles, and \ours{}, all with HT weights and the
same benign budget. Cutoffs and budgets are fixed before evaluating
$500$ draws per configuration. Each draw's mean of five per-seed
metrics is compared with the full-test mean
(\Cref{app:heldout_sampling}).

\begin{table}[t]
\centering\small
\setlength{\tabcolsep}{3pt}
\begin{tabular}{lrrr}
\toprule
\textbf{Sampler} & \shortstack{\textbf{F1 RMSE}\\(pp)} & \shortstack{\textbf{F1 CI width}\\(pp)} & \shortstack{\textbf{Coverage}\\(\%)} \\
\midrule
\multicolumn{4}{l}{\textit{HI-Small: LGB $\to$ XGB}} \\
Uniform + HT & 6.76 & 40.3 & 100.0 \\
Terciles + HT & 6.74 & 58.7 & 99.6 \\
HT-Coreset & \textbf{0.00} & 55.0 & 100.0 \\
\addlinespace[2pt]
\multicolumn{4}{l}{\textit{HI-Small: XGB $\to$ LGB}} \\
Uniform + HT & 7.41 & 40.4 & 99.8 \\
Terciles + HT & 7.60 & 58.3 & 99.6 \\
HT-Coreset & \textbf{1.59} & 53.2 & 100.0 \\
\addlinespace[2pt]
\multicolumn{4}{l}{\textit{LI-Small: LGB $\to$ XGB}} \\
Uniform + HT & 4.44 & 26.5 & 100.0 \\
Terciles + HT & 4.53 & 29.7 & 100.0 \\
HT-Coreset & \textbf{0.64} & 27.5 & 100.0 \\
\addlinespace[2pt]
\multicolumn{4}{l}{\textit{LI-Small: XGB $\to$ LGB}} \\
Uniform + HT & 6.02 & 27.1 & 99.8 \\
Terciles + HT & 5.76 & 30.4 & 100.0 \\
HT-Coreset & \textbf{2.58} & 27.3 & 100.0 \\
\addlinespace[2pt]
\bottomrule\end{tabular}
\caption{Sampling with a held-out predictor family at the released target counts.
LGB=LightGBM; XGB=XGBoost. The arrow runs from the stratum constructor to the evaluated family.
All samplers retain every illicit edge and use HT weights.
HT-Coreset combines hard-negative retention with difficulty terciles.
RMSE and mean $95\%$ confidence-interval (CI) width concern F1 and are in percentage points;
coverage is a percentage over $500$ draws.
Each draw estimates the mean of five fixed, validation-thresholded seed metrics.
Bold marks the lowest F1 RMSE within each dataset--direction block.
Intervals are conservative and can remain broad despite small point errors.}
\label{tab:heldout_sampling}
\end{table}

At the released budgets, hard-negative sampling gives F1 RMSE
of $0.00$--$2.58$ pp, versus $4.44$--$7.41$ pp for uniform benign
sampling (Table~\ref{tab:heldout_sampling}). It has the lowest
F1 RMSE in all $12$ dataset--direction--budget comparisons.
At matched budgets, concentrating samples on likely false positives
reduces point error for these two families. Their estimates still
vary when they flag benign edges outside the constructor's hard stratum.
Conservative intervals cover $99.6$--$100\%$ of full-test F1 values
across configurations, but mean widths at the released budgets
remain $27.3$--$55.0$ pp. Thus, the point estimates improve without
establishing precise estimation for every predictor.

\section{LLM Evaluation}
\label{sec:llm}

The retained targets make repeated model and prompt comparisons
possible with thousands of requests per pass. We ask whether the
LLMs distinguish illicit from benign transactions and identify
illicit typologies. Detection is assessed against a predict-all-illicit
reference, which achieves high recall without distinguishing cases.

\subsection{Evaluation Setup}
\label{sec:setup}

All results here use the original typed-graph prompts described
in \Cref{sec:method:serial}.

\paragraph{ML reference.}
The two GFP boosters let us compare which transactions ML and
LLMs misclassify. They receive GFP features, while LLMs receive
graph text. We average LightGBM+GFP and XGBoost+GFP probabilities
over five training seeds and apply operating thresholds of $0.80$
on \hiSmall and $0.48$ on \liSmall. These thresholds were selected
on full-test labels, so this is an exploratory comparison.
GCPAL enters only the frozen sampling design; results and the
three-to-two-model comparison appear in \Cref{app:non_llm}.

\paragraph{LLMs.}
We evaluate $7$ open-weight thinking models spanning $3$--$27$B
active parameters across $3$ families:
GPT-OSS-\{20B, 120B\}, Nemotron-3-\{Nano-30B, Super-120B\},
Qwen3.5-\{35B-A3B, 397B-A17B, 27B\}. Model configurations and
sampling parameters are in \Cref{tab:sampling_params}.
Qwen3.5-27B's deployed $128$K context
cap causes input-plus-output budget rejections
(\Cref{app:examples}).

\paragraph{Prompting variants.}
To test the effect of demonstrations, we keep task instructions
and typed $k$-hop graph serialization fixed.
\textbf{\fs} (few-shot, ``FS'') prepends $12$ training-set
demonstrations, following \citet{pirmorad2025iclaml}: one illicit
example per typology ($8$ total) and $4$ benign examples.
\textbf{\zs} (zero-shot, ``ZS'') uses the same instructions and
case graph without demonstrations.

\paragraph{Repeated runs and scoring.}
Each variant uses $5$ stochastic runs with fixed targets and,
for \fs, a fixed demonstration pool. Stored seed identifiers
were not passed to the model API (\Cref{app:seed_std}).
Detection metrics include all targets; the archived binary scorer
defaults missing predictions to benign. The separate error taxonomy
records no response as its own outcome. Typology macro-F1 (TF1)
uses illicit edges with a reference typology, counting missing or
unparsed labels as their own label (\Cref{app:llm_results_group}).
The recorded-token ratio ($V\%$) measures whether any tokens were
returned, not whether a response was parseable.
Unweighted detection F1 run SD is below $0.5$ pp in $26$ of $28$
model--method--split cells (\Cref{tab:llm_ht_full}).

\subsection{Detection and Typology Results}
\label{sec:llm:results}

\textbf{HT-weighted} F1 estimates detection on the full test split;
\textbf{Lift} is HT precision divided by that split's illicit
prevalence. Lift $1$ is the predict-all reference; larger values
indicate greater precision. \textbf{Unweighted} F1 and precision
(P) describe performance on the enriched $1{:}2$ subset.
Recall (R) measures recovered illicit edges; TF1 measures typology
classification. All four are percentages.

\begin{table}[t]
\centering
\small
\setlength{\tabcolsep}{1.25pt}
\begin{tabular}{@{}llcccccc@{}}
\toprule
& & \multicolumn{2}{c}{\footnotesize\shortstack{\textbf{HT-weighted}\\(full test)}} & \multicolumn{2}{c}{\footnotesize\shortstack{\textbf{Unweighted}\\(subset)}} & & \\
\cmidrule(lr){3-4}\cmidrule(lr){5-6}
\textbf{Model} & \textbf{M} & \textbf{F1} & \textbf{Lift} & F1 & P & \textbf{R} & \textbf{TF1} \\
\midrule
GPT-OSS-20B       & FS & 0.236 & 0.96 & 46.4 & 32.8 & 79.2 & 10.5 \\
GPT-OSS-20B       & ZS & 0.246 & 1.00 & 49.9 & 33.3 & 99.5 & 5.6  \\
Qwen3.5-27B       & FS & 0.240 & 0.98 & 40.7 & 32.5 & 54.5 & 10.1 \\
Qwen3.5-27B       & ZS & 0.230 & 0.94 & 45.8 & 31.6 & 82.9 & 6.3  \\
Nem-Nano-30B      & FS & 0.245 & 1.00 & 49.5 & 33.4 & 95.4 & 7.5  \\
Nem-Nano-30B      & ZS & 0.246 & 1.00 & 50.0 & 33.3 & 99.9 & 6.6  \\
Qwen3.5-35B       & FS & 0.253 & 1.03 & 48.5 & 34.0 & 84.9 & 16.5 \\
Qwen3.5-35B       & ZS & 0.246 & 1.00 & 49.7 & 33.3 & 97.8 & 6.1  \\
GPT-OSS-120B      & FS & 0.249 & 1.01 & 50.3 & 34.0 & 96.4 & 10.1 \\
GPT-OSS-120B      & ZS & 0.246 & 1.00 & 49.9 & 33.3 & 99.8 & 8.1  \\
Nem-Super-120B    & FS & 0.247 & 1.01 & 47.9 & 33.7 & 82.4 & 12.3 \\
Nem-Super-120B    & ZS & 0.247 & 1.00 & 50.0 & 33.5 & 99.0 & 7.8  \\
Qwen3.5-397B      & FS & 0.251 & 1.02 & 50.6 & 34.1 & 98.3 & \textbf{18.4} \\
Qwen3.5-397B      & ZS & 0.248 & 1.01 & 50.2 & 33.5 & 100.0 & 7.6  \\
\midrule
Predict-all illicit & --- & 0.246 & 1.00 & 50.0 & 33.3 & 100.0 & --- \\
\bottomrule
\end{tabular}
\caption{LLM results on \hiSmall, means over $5$ runs.
All metrics are percentages except Lift; M is prompting method
(FS=\fs, ZS=\zs). Both summaries use the same HT-Coreset targets and predictions.
Qwen3.5-35B denotes the 35B-A3B model.
R and TF1 are unchanged by weighting because their illicit edges
have weight one. Full results for both splits and unweighted run
SD are in \Cref{tab:llm_ht_full}; HT run SD and sampling uncertainty
are in \Cref{tab:llm_sampling_uncertainty}.}
\label{tab:llm_results}
\end{table}

\paragraph{High recall accompanies extensive false positives.}
On \hiSmall, $F1_{\text{HT}}$ ranges from $0.230\%$ to
$0.253\%$, near the predict-all-illicit reference of $0.246\%$;
precision lift is $0.94$--$1.03$. Unweighted precision is
$31.6$--$34.1\%$, close to the subset's $33.3\%$ illicit proportion.
Zero-shot recall reaches $97.8$--$100\%$ for six models and
$82.9\%$ for Qwen3.5-27B. Thus, high recall here provides little
selectivity: many benign transactions are also flagged.
Unweighted F1 is $40.7$--$50.6\%$, below the ML reference's
$68.1\%$.
Cellwise $95\%$ precision-lift intervals, based on observed benign
predictions within each stratum, have maximum upper endpoints of
$1.086$ on \hiSmall and $1.051$ on \liSmall; some lower endpoints
lie slightly above one (\Cref{app:llm_sampling}). These estimates
condition on the frozen design and five fixed prediction vectors;
closeness to the baseline does not establish statistical equivalence.

\paragraph{Larger models do not consistently improve these scores.}
Excluding Qwen3.5-27B's context-limited deployment,
$F1_{\text{HT}}$ varies by $0.017$ pp across a ${\sim}5.7\times$
active-parameter range. The $3$B-active Qwen3.5-35B-A3B reaches
$16.5\%$ TF1, within $1.9$ pp of the $17$B-active Qwen3.5-397B
($18.4\%$), the highest score in the table. These are comparisons
of deployed models, not a controlled test of parameter scaling.

\paragraph{Demonstrations help typology identification more consistently.}
Few-shot TF1 exceeds zero-shot TF1 for all seven models on \hiSmall
and six on \liSmall (\Cref{app:fewshot}); the mean \hiSmall gain
is $5.3$ pp. Detection differences depend on the evaluation frame:
mean unweighted F1 falls by $1.6$ pp ($5/7$ models decrease), while
$F1_{\text{HT}}$ rises by $0.002$ pp ($5/7$ increase).
The prompts therefore improve typology scores without consistently
improving detection on the selected cases.

\section{Diagnosing LLM Outputs}
\label{sec:llm:why}

The detection results motivate two questions: do LLMs recover cases
missed by ML, and do observable trace markers help distinguish correct
answers? Error decompositions and trace analyses use the available
records from the run labeled $42$; the label-concentration figure
pools all five runs. Unless marked HT, counts and rates describe
the selected cases or the stated audit sample. Written traces need
not faithfully represent internal computation
\citep{turpin2023language,tutek2025measuring}.

\subsection{Decomposing ML and LLM Errors}
\label{sec:llm:asymmetry}

The \hiSmall ML reference has $84.8\%$ precision and $56.8\%$
recall. Most zero-shot LLMs have much higher recall but flag nearly
three times the true illicit count (\Cref{app:over_predict_stats}).
To examine complementarity, Table~\ref{tab:transitions} groups edges
by whether ML, GPT-OSS-120B, both, or neither gives the correct
binary verdict. Ground-truth labels determine these groups.

Of $2{,}123$ ML-only-correct edges, $2{,}064$ are benign; the group's
mean ML score is $0.08$. Of $433$ LLM-only-correct edges, $410$
are illicit; the group's mean ML score is $0.13$. Both means are far
below the $0.80$ ML threshold. The LLM recovers missed illicit
cases but also flags benign cases with low ML scores. A useful
router would have to distinguish these groups using observable
information; their ground-truth-defined means do not give such a
rule. Full transitions and per-typology results are in
\Cref{app:error_dist,app:error_transitions}.

\begin{table}[t]
\centering
\small
\setlength{\tabcolsep}{3.5pt}
\begin{tabular}{lrrr}
\toprule
\textbf{Detection outcome} & \textbf{Illicit} & \textbf{Benign} & $\bar{s}_{\mathrm{ML}}$ \\
\midrule
Both correct & 652 & 311 & 0.69 \\
ML only correct & 59 & 2{,}064 & 0.08 \\
LLM only correct & 410 & 23 & 0.13 \\
Both wrong & 130 & 104 & 0.40 \\
\bottomrule
\end{tabular}
\caption{Detection-error decomposition on \hiSmall
(GPT-OSS-120B, few-shot; $3{,}753$ edges).
Illicit and Benign are ground-truth edge counts.
Groups use binary detection correctness. Each edge's ML score averages
the two boosters' illicit probabilities over five training seeds.
$\bar{s}_{\mathrm{ML}}$ averages these scores within each group;
ML decisions use threshold $0.80$.}
\label{tab:transitions}
\end{table}

\subsection{Predicted Labels and Trace Length}
\label{sec:llm:traces}

\begin{figure*}[!t]
\centering
\includegraphics[width=\textwidth]{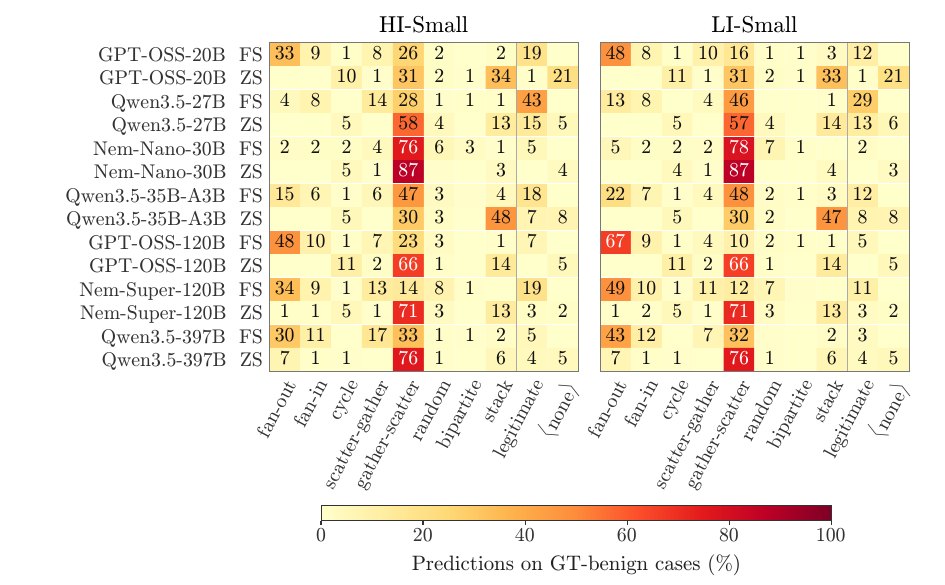}
\caption{Predicted labels on ground-truth-benign cases in
\hiSmall (HI) and \liSmall (LI). Each row shows percentages of
all benign cases pooled over five runs. Legitimate is the benign
label; $\langle\mathrm{none}\rangle$ denotes a missing typology label.
Typology concentrations vary with model and prompting; dominant-label
shares appear in \Cref{app:typology_freq}.}
\label{fig:anchor_heatmap}
\end{figure*}

With no response treated as a separate outcome, false positives
account for $38.7$--$63.6\%$ of \hiSmall few-shot cases across
models, while false negatives account for at most $4.7\%$
(\Cref{tab:failure_taxonomy}). Qwen3.5-27B returns no response on
$38.7\%$ of cases, mostly because of its deployed context budget
(\Cref{app:examples}). Figure~\ref{fig:anchor_heatmap} shows which
labels models assign to ground-truth-benign cases. GPT-OSS-120B
favors fan-out under few-shot prompting and gather-scatter under
zero-shot prompting; Nemotron-Nano favors gather-scatter in both.
These concentrations identify model- and prompt-specific false-positive
patterns to inspect, without establishing why a model chose a label
(\Cref{app:typology_freq}).

Few-shot traces are longer on average than zero-shot traces for
every model in this comparison. Within models, however, mean lengths
for correct and incorrect verdicts have no consistent ordering.
These aggregate comparisons do not support using length as a
confidence measure; full lengths and conditional differences are
reported in \Cref{app:trace_stats}.

\subsection{Lexical Checks and Answer Correctness}
\label{sec:llm:failure}

We test whether traces that name structural features, candidate
typologies, and supporting markers are more likely to accompany a
correct verdict. The audit contains $1{,}000$ outcome-stratified
\hiSmall few-shot records across seven models, including $148$
empty traces. The original rubric, motivated by step-level process
supervision~\citep{lightman2024lets}, checks \textbf{Parse}
(features), \textbf{Recall} (typologies), \textbf{Match} (structural
markers), and \textbf{Conclude} (reference verdict and, when illicit,
typology). Regex rules and three API judges score this sample
(\Cref{app:multi_judge}).

For the correctness comparison, revised regex checks require at
least two feature categories, two typology-name strings, and two
marker categories for every verdict. This removes the original
Match rule's automatic pass for nonempty benign verdicts.
We repeat the checks on the first $8{,}000$ characters to match the
judges' text window, using the same frozen model outputs.

\begin{table}[!ht]
\centering\small
\setlength{\tabcolsep}{4pt}
\begin{tabular}{lrrr}
\toprule
\textbf{Binary verdict} & $n$ & \textbf{Full trace} & \textbf{First 8k chars} \\
\midrule
Correct & 470 & 397 (84.5\%) & 388 (82.6\%) \\
Incorrect & 530 & 298 (56.2\%) & 295 (55.7\%) \\
\bottomrule\end{tabular}
\caption{Revised lexical checks by binary correctness in the $1{,}000$-trace audit.
Counts require Parse, Recall, and at least two structural-marker categories for every
verdict, removing the original benign-verdict Match bypass. Percentages use the row
denominator. The 8k-character window matches the API judges' text limit. These are
outcome-stratified descriptive rates, not population estimates.}
\label{tab:audit_sensitivity}
\end{table}

The checks pass for $84.5\%$ of binary-correct records ($397/470$)
and $56.2\%$ of binary-incorrect records ($298/530$).
With the shorter window, these rates are $82.6\%$ and $55.7\%$
(Table~\ref{tab:audit_sensitivity}). Thus, the markers are more
common in correct outputs but also occur in many incorrect ones;
passing them is insufficient for correctness. The checks detect
written markers rather than verify that graph evidence supports
them. Scoring-rule sensitivity, typology-inclusive correctness,
and per-model counts appear in \Cref{app:audit_sensitivity}.

Three human raters assess a separate $28$-record sample, one record
per model in each of four outcome groups. Match agreement is low
($\kappa=0.10$); Conclude uses a shared parser rule. The API judges
also receive no transaction graph. These checks assess rubric
application within the samples and cannot independently validate
graph grounding or estimate population failure rates
(\Cref{app:human_rating,app:audit_sensitivity}).

\subsection{A Supplementary Routing Diagnostic}
\label{sec:llm:routing}

The error overlap motivates testing how much a hybrid comparison's
gain comes from the LLM versus the selected route. Doubt Triage (DT)
consults an LLM on ML-negative edges in the weight-one stratum
(all illicit edges and retained hard-negative benign edges) and
uses ML elsewhere. Its best zero-shot detection F1, averaged over
five runs, is $87.3\%$ on \hiSmall (ML: $68.1\%$) and $84.9\%$
on \liSmall (ML: $28.6\%$). An always-illicit answer on the same
route, without LLM calls, also reaches $87.3\%$ and $84.9\%$.
The route therefore accounts for most of the apparent gain.
Because it uses label-derived strata, DT does not establish a
rule for selecting cases at deployment. \Cref{sec:hybrid} reports
the control and tested score-only routes.

\section{Discussion and Conclusion}
\label{sec:discussion}

\bench{} enables model and prompt comparisons with thousands
of evaluation targets while retaining every illicit test case.
To reuse it, align predictions to the released case IDs and apply
the supplied inclusion weights to confusion counts before computing
precision and F1. Report HT-weighted estimates for full-test
performance and unweighted errors for the selected cases, stating
which input version was evaluated. A new labeled transaction
graph requires its own strata and inclusion probabilities.

The construction checks show metric agreement for the ensemble
used to select targets. Holding out a GFP booster family gives
lower point error than uniform benign sampling at matched budgets,
although conservative intervals remain broad. This supports the
benign allocation for the tested families while leaving transfer
to other predictors open.

With the original graph prompts, the evaluated LLMs flag many
benign transactions; high recall accompanies precision near
full-test illicit prevalence. Demonstration gains are more consistent
for typology identification than for detection. Prompt comparisons
should report these tasks separately. In the audit sample, lexical
checks pass more often for correct answers and also for over
half of incorrect binary verdicts. Trace audits should compare
check outcomes with answer correctness rather than treat a pass as
evidence that the answer is supported by the graph.
A paired comparison of original and target-marked inputs, holding
models, demonstrations, and scoring fixed, could measure sensitivity
to the combined input changes. The supplementary frontier probe
uses a different protocol and does not provide that comparison
(\Cref{app:probe}).

\section*{Limitations}
\label{sec:limitations}

The main LLM evaluation covers seven open-weight models on two
small synthetic AMLworld splits with one graph-to-text pipeline.
Generalizability to larger graphs and real banking data remains
untested. The evaluated inputs do not consistently identify and
include the target transaction (\Cref{sec:method:serial}), so their
errors cannot be attributed solely to model reasoning.

The CSV-order partitions have overlapping timestamps and do not
establish strictly prospective ML performance. The task ML reference
and DT use thresholds selected on full-test labels, making comparisons
at these thresholds exploratory.

LLMs were evaluated at one coreset size and one $50$-neighbor
serialization cap. Sampling uncertainty conditions on the frozen
design and predictions; it has not been checked against full-test
LLM predictions or independently evaluated coreset draws
(\Cref{app:estimation}). The held-out-family study tests two related
GFP boosters and does not establish transfer to unrelated predictors.

Trace-level findings use the available run, and their stability
across runs is untested. The $28$-record human check assesses
rubric application on an outcome-stratified sample
(\Cref{app:human_rating}).

\section*{Ethical Considerations}
\label{sec:ethics}

AMLworld~\citep{altman2023realistic} is synthetic: its account
identifiers, amounts, and relationships do not expose real customers.
The supplementary Elliptic experiment uses public Bitcoin transactions
(\Cref{app:elliptic}). The intended use is better evaluation of
AML detection, but examples of laundering patterns could also help
attempts to evade detection. False positives can burden legitimate
customers and compliance teams; the reported over-prediction does
not support LLM-only deployment. DT uses label-informed evaluation
strata, and retaining ML in the loop does not establish deployment
safety. Release licenses and the code-only Elliptic distribution
are documented in \Cref{app:licenses}.

\paragraph{AI Usage Statement.}
The authors used AI assistants (Claude and GPT-class models) for
manuscript editing and code review. The authors are responsible
for the experimental design, interpretation, and final manuscript.

\makeatletter
\ifacl@anonymize\else
\section*{Acknowledgments}

This work began at SCB 10X. We thank
\href{https://th.linkedin.com/in/ornsmith}{Oravee Smithiphol} for introducing
our team to Phume Ngampornsukswadi at Scotiabank.
We also thank Duncan Halverson for his feedback on an earlier draft and the
members of the \href{https://opentyphoon.ai/}{Typhoon team}.
\fi
\makeatother

\bibliography{references}

@inproceedings{altman2023realistic,
  title={Realistic Synthetic Financial Transactions for Anti-Money Laundering Models},
  author={Altman, Erik and Blanu{\v{s}}a, Jovan and von Niederh{\"a}usern, Luc and Egressy, B{\'e}ni and Anghel, Andreea and Atasu, Kubilay},
  booktitle={Thirty-seventh Conference on Neural Information Processing Systems Datasets and Benchmarks Track},
  year={2023},
  url={https://openreview.net/forum?id=XZf2bnMBag}
}

@inproceedings{pirmorad2025iclaml,
  title={Exploring the In-Context Learning Capabilities of {LLM}s for Money Laundering Detection in Financial Graphs},
  author={Pirmorad, Erfan},
  booktitle={2025 IEEE International Conference on Data Mining Workshops (ICDMW)},
  year={2025},
  note={AI4FCF workshop; arXiv:2507.14785},
  doi={10.1109/ICDMW69685.2025.00026},
  url={https://arxiv.org/abs/2507.14785}
}

@article{lu2024gcpal,
  title={Graph Contrastive Pre-training for Anti-money Laundering},
  author={Lu, Hanbin and Wang, Haosen},
  journal={International Journal of Computational Intelligence Systems},
  volume={17},
  number={1},
  pages={307},
  year={2024},
  publisher={Springer},
  doi={10.1007/s44196-024-00720-4},
  url={https://link.springer.com/article/10.1007/s44196-024-00720-4}
}

@inproceedings{xu2019powerful,
  title={How Powerful Are Graph Neural Networks?},
  author={Xu, Keyulu and Hu, Weihua and Leskovec, Jure and Jegelka, Stefanie},
  booktitle={International Conference on Learning Representations},
  year={2019},
  url={https://openreview.net/forum?id=ryGs6iA5Km}
}

@article{fleiss1971agreement,
  title={Measuring Nominal Scale Agreement among Many Raters},
  author={Fleiss, Joseph L.},
  journal={Psychological Bulletin},
  volume={76},
  number={5},
  pages={378--382},
  year={1971},
  doi={10.1037/h0031619},
  url={https://doi.org/10.1037/h0031619}
}

@article{cohen1960agreement,
  title={A Coefficient of Agreement for Nominal Scales},
  author={Cohen, Jacob},
  journal={Educational and Psychological Measurement},
  volume={20},
  number={1},
  pages={37--46},
  year={1960},
  doi={10.1177/001316446002000104},
  url={https://doi.org/10.1177/001316446002000104}
}

@article{horvitz1952generalization,
  title={A Generalization of Sampling Without Replacement From a Finite Universe},
  author={Horvitz, Daniel G and Thompson, Donovan J},
  journal={Journal of the American Statistical Association},
  volume={47},
  number={260},
  pages={663--685},
  year={1952},
  url={https://doi.org/10.1080/01621459.1952.10483446}
}

@book{sarndal1992model,
  title={Model Assisted Survey Sampling},
  author={S{\"a}rndal, Carl-Erik and Swensson, Bengt and Wretman, Jan},
  year={1992},
  publisher={Springer},
  address={New York},
  doi={10.1007/978-1-4612-4378-6}
}

@article{neyman1934allocation,
  title={On the Two Different Aspects of the Representative Method: The Method of Stratified Sampling and the Method of Purposive Selection},
  author={Neyman, Jerzy},
  journal={Journal of the Royal Statistical Society},
  volume={97},
  number={4},
  pages={558--625},
  year={1934},
  url={https://doi.org/10.2307/2342192}
}

@inproceedings{polo2024tinybenchmarks,
  title={tiny{B}enchmarks: Evaluating {LLM}s with Fewer Examples},
  author={Maia Polo, Felipe and Weber, Lucas and Choshen, Leshem and Sun, Yuekai and Xu, Gongjun and Yurochkin, Mikhail},
  booktitle={Proceedings of the 41st International Conference on Machine Learning},
  series={Proceedings of Machine Learning Research},
  volume={235},
  pages={34303--34326},
  year={2024},
  publisher={PMLR},
  url={https://proceedings.mlr.press/v235/maia-polo24a.html}
}

@inproceedings{kipnis2025metabench,
  title={metabench -- A Sparse Benchmark of Reasoning and Knowledge in Large Language Models},
  author={Kipnis, Alex and Voudouris, Konstantinos and Schulze Buschoff, Luca M and Schulz, Eric},
  booktitle={ICLR},
  year={2025},
  url={https://arxiv.org/abs/2407.12844}
}

@inproceedings{fogliato2024reducedlabel,
  title={A Framework for Efficient Model Evaluation through Stratification, Sampling, and Estimation},
  author={Fogliato, Riccardo and Patil, Pratik and Monfort, Mathew and Perona, Pietro},
  booktitle={ECCV},
  year={2024},
  url={https://arxiv.org/abs/2406.07320}
}

@inproceedings{mirzasoleiman2020coresets,
  title={Coresets for Data-efficient Training of Machine Learning Models},
  author={Mirzasoleiman, Baharan and Bilmes, Jeff and Leskovec, Jure},
  booktitle={Proceedings of the 37th International Conference on Machine Learning (ICML)},
  series={Proceedings of Machine Learning Research},
  volume={119},
  pages={6950--6960},
  year={2020},
  publisher={PMLR},
  url={https://proceedings.mlr.press/v119/mirzasoleiman20a.html}
}

@inproceedings{leskovec2006sampling,
  title={Sampling from Large Graphs},
  author={Leskovec, Jure and Faloutsos, Christos},
  booktitle={Proceedings of the 12th ACM SIGKDD International Conference on Knowledge Discovery and Data Mining (KDD)},
  pages={631--636},
  year={2006},
  url={https://dl.acm.org/doi/10.1145/1150402.1150479}
}

@article{gao2019efficient,
  title={Efficient Knowledge Graph Accuracy Evaluation},
  author={Gao, Junyang and Li, Xian and Xu, Yifan Ethan and Sisman, Bunyamin and Dong, Xin Luna and Yang, Jun},
  journal={Proceedings of the VLDB Endowment},
  volume={12},
  number={11},
  pages={1679--1691},
  year={2019},
  url={https://doi.org/10.14778/3342263.3342642}
}

@inproceedings{chen2016xgboost,
  title={{XGB}oost: A Scalable Tree Boosting System},
  author={Chen, Tianqi and Guestrin, Carlos},
  booktitle={Proceedings of the 22nd ACM SIGKDD International Conference on Knowledge Discovery and Data Mining (KDD)},
  pages={785--794},
  year={2016},
  publisher={ACM},
  doi={10.1145/2939672.2939785},
  url={https://doi.org/10.1145/2939672.2939785}
}

@inproceedings{ke2017lightgbm,
  title={Light{GBM}: A Highly Efficient Gradient Boosting Decision Tree},
  author={Ke, Guolin and Meng, Qi and Finley, Thomas and Wang, Taifeng and Chen, Wei and Ma, Weidong and Ye, Qiwei and Liu, Tie-Yan},
  booktitle={Advances in Neural Information Processing Systems (NeurIPS)},
  volume={30},
  year={2017},
  publisher={Curran Associates, Inc.},
  url={https://proceedings.neurips.cc/paper/2017/hash/6449f44a102fde848669bdd9eb6b76fa-Abstract.html}
}

@article{chen2024llmgraph,
  title={Exploring the Potential of Large Language Models ({LLM}s) in Learning on Graphs},
  author={Chen, Zhikai and Mao, Haitao and Li, Hang and Jin, Wei and Wen, Hongzhi and Wei, Xiaochi and Wang, Shuaiqiang and Yin, Dawei and Fan, Wenqi and Liu, Hui and Tang, Jiliang},
  journal={ACM SIGKDD Explorations Newsletter},
  volume={25},
  number={2},
  year={2024},
  url={https://arxiv.org/abs/2307.03393}
}

@inproceedings{fatemi2024talk,
  title={Talk like a Graph: Encoding Graphs for Large Language Models},
  author={Fatemi, Bahare and Halcrow, Jonathan and Perozzi, Bryan},
  booktitle={The Twelfth International Conference on Learning Representations (ICLR)},
  year={2024},
  url={https://openreview.net/forum?id=IuXR1CCrSi}
}

@inproceedings{lightman2024lets,
  title={Let's Verify Step by Step},
  author={Lightman, Hunter and Kosaraju, Vineet and Burda, Yuri and Edwards, Harrison and Baker, Bowen and Lee, Teddy and Leike, Jan and Schulman, John and Sutskever, Ilya and Cobbe, Karl},
  booktitle={The Twelfth International Conference on Learning Representations (ICLR)},
  year={2024},
  url={https://openreview.net/forum?id=v8L0pN6EOi}
}

@inproceedings{weber2019anti,
  title={Anti-Money Laundering in Bitcoin: Experimenting with Graph Convolutional Networks for Financial Forensics},
  author={Weber, Mark and Domeniconi, Giacomo and Chen, Jie and Weidele, Daniel Karl I. and Bellei, Claudio and Robinson, Tom and Leiserson, Charles E.},
  booktitle={Anomaly Detection in Finance Workshop at the 25th ACM SIGKDD Conference on Knowledge Discovery and Data Mining (KDD)},
  year={2019},
  url={https://arxiv.org/abs/1908.02591}
}

@article{edge2024graphrag,
  title={From Local to Global: A {Graph RAG} Approach to Query-Focused Summarization},
  author={Edge, Darren and Trinh, Ha and Cheng, Newman and Bradley, Joshua and Chao, Alex and Mody, Apurva and Truitt, Steven and Metropolitansky, Dasha and Ness, Robert Osazuwa and Larson, Jonathan},
  journal={arXiv preprint arXiv:2404.16130},
  year={2024},
  url={https://arxiv.org/abs/2404.16130}
}

@inproceedings{turpin2023language,
  title={Language Models Don't Always Say What They Think: Unfaithful Explanations in Chain-of-Thought Prompting},
  author={Turpin, Miles and Michael, Julian and Perez, Ethan and Bowman, Samuel R.},
  booktitle={NeurIPS},
  year={2023},
  url={https://openreview.net/forum?id=bzs4uPLXvi}
}

@inproceedings{tutek2025measuring,
  title={Measuring Chain of Thought Faithfulness by Unlearning Reasoning Steps},
  author={Tutek, Martin and Hashemi Chaleshtori, Fateme and Marasovi{\'c}, Ana and Belinkov, Yonatan},
  booktitle={EMNLP},
  pages={9935--9960},
  year={2025},
  url={https://aclanthology.org/2025.emnlp-main.504/},
  doi={10.18653/v1/2025.emnlp-main.504}
}

@article{landis1977measurement,
  title={The Measurement of Observer Agreement for Categorical Data},
  author={Landis, J. Richard and Koch, Gary G.},
  journal={Biometrics},
  volume={33},
  number={1},
  pages={159--174},
  year={1977},
  doi={10.2307/2529310}
}

@inproceedings{blanusa2024gfp,
  title={Graph Feature Preprocessor: Real-time Subgraph-based Feature Extraction for Financial Crime Detection},
  author={Blanu{\v{s}}a, Jovan and Cravero Baraja, Maximo and Anghel, Andreea and von Niederh{\"a}usern, Luc and Altman, Erik and Pozidis, Haris and Atasu, Kubilay},
  booktitle={Proceedings of the 5th ACM International Conference on AI in Finance (ICAIF '24)},
  pages={222--230},
  year={2024},
  doi={10.1145/3677052.3698674},
  url={https://doi.org/10.1145/3677052.3698674}
}

@inproceedings{zheng2023coveragecentric,
  title={Coverage-centric Coreset Selection for High Pruning Rates},
  author={Haizhong Zheng and Rui Liu and Fan Lai and Atul Prakash},
  booktitle={The Eleventh International Conference on Learning Representations},
  year={2023},
  url={https://openreview.net/forum?id=QwKvL6wC8Yi}
}

\appendix
\crefalias{section}{appendix}
\crefalias{subsection}{appendix}
\crefalias{subsubsection}{appendix}
\crefalias{subappendix}{appendix}
\crefalias{subsubappendix}{appendix}
\raggedbottom
\makeatletter
\setlength{\@fptop}{0pt}
\setlength{\@fpsep}{12pt}
\setlength{\@dblfptop}{0pt}
\setlength{\@dblfpsep}{12pt}
\newcommand{\appendixkeepheading}[1]{%
  \par\begingroup
  \@tempdima=#1\relax
  \advance\@tempdima by \pagetotal
  \ifdim\@tempdima>\pagegoal\newpage\fi
  \endgroup}
\makeatother

\section{Dataset Details}
\label{app:dataset}

The complete \hiSmall and \liSmall source datasets have illicit
rates of $0.10\%$ and $0.05\%$, respectively
(\citealp[Appendix~A, Table~4]{altman2023realistic}). These are
full-dataset rates; Table~\ref{tab:resource_summary} reports the
test-partition rates and released-set sizes used here.

\paragraph{Typology Definitions.}
\label{app:typologies}
The 8 AML typologies in AMLworld~\citep[\S3.2, Fig.~2]{altman2023realistic} are:
\begin{enumerate}
\item \textbf{Fan-out:} One source distributes to many recipients
\item \textbf{Fan-in:} Many sources aggregate to one recipient
\item \textbf{Cycle:} Circular fund flow returning to the origin
\item \textbf{Scatter-gather:} Funds scattered then regathered
\item \textbf{Gather-scatter:} Funds gathered then scattered
\item \textbf{Stack:} Layered sequential transfers
\item \textbf{Bipartite:} Two-group transfer pattern
\item \textbf{Random:} Random-walk transfers among controlled accounts without returning to the origin
\end{enumerate}

\paragraph{Release licenses.}
\label{app:licenses}
Our code, released ML parameters, and full-test model outputs are
licensed under MIT. The evaluation dataset and derived coreset
artifacts are released under the Community Data License Agreement,
Sharing, Version 1.0, as specified in the release notice.
The Elliptic experiments are released as code only: Elliptic is
CC BY-NC-ND 4.0, whose NoDerivatives clause forbids redistributing
adapted material, so we ship the pipeline that reproduces them from
the original release rather than any derived data.

\paragraph{Ground-truth typology prior in \ours subsets.}
Table~\ref{tab:typology_prior} reports the per-typology illicit-edge
counts in the \ours{} subset, providing class supports for the
per-typology analyses.

\begin{table}[!htbp]
\centering
\small
\setlength{\tabcolsep}{4pt}
\begin{tabular}{lrrrr}
\toprule
\textbf{Typology} & \multicolumn{2}{c}{\textbf{HI-Small}} & \multicolumn{2}{c}{\textbf{LI-Small}} \\
\cmidrule(lr){2-3}\cmidrule(lr){4-5}
& $n$ & \% & $n$ & \% \\
\midrule
gather-scatter   & $259$ & $32.7$ &  $27$ & $15.5$ \\
scatter-gather   & $142$ & $18.0$ &  $30$ & $17.2$ \\
fan-in           & $\phantom{0}90$ & $11.4$ &  $11$ & $\phantom{0}6.3$ \\
fan-out          & $\phantom{0}85$ & $10.7$ &  $29$ & $16.7$ \\
cycle            & $\phantom{0}74$ & $\phantom{0}9.4$ &  $30$ & $17.2$ \\
random           & $\phantom{0}66$ & $\phantom{0}8.3$ &  $10$ & $\phantom{0}5.7$ \\
bipartite        & $\phantom{0}43$ & $\phantom{0}5.4$ &  $27$ & $15.5$ \\
stack            & $\phantom{0}32$ & $\phantom{0}4.0$ &  $10$ & $\phantom{0}5.7$ \\
\midrule
Labeled illicit  & $791$ & $100.0$ & $174$ & $100.0$ \\
\bottomrule
\end{tabular}
\caption{Ground-truth typology distribution in \ours subsets
(typology-labeled illicit edges; \% within labeled-illicit). Of the
illicit edges, $791$ of $1{,}251$ on \hiSmall and $174$ of $756$ on
\liSmall carry explicit typology labels in AMLworld; the remainder are illicit
without a programmatic typology assignment. \emph{stack} is the
rarest labeled typology ($4.0\%$ \hiSmall), \emph{gather-scatter} the
most common ($32.7\%$ \hiSmall).}
\label{tab:typology_prior}
\end{table}

\appendixkeepheading{7\baselineskip}
\section{Graph Inputs and Target Identification}
\label{app:subgraph}

Table~\ref{tab:subgraph_params} reports our subgraph extraction
settings alongside the closest prior work. Neither
\citet{pirmorad2025iclaml} (which introduced the typed-graph
protocol we adopt) nor \citet{altman2023realistic}'s GNN baseline
specifies an LLM-prompting neighbor cap, so we set both values
explicitly.

\begin{table}[!htbp]
\centering
\small
\setlength{\tabcolsep}{3pt}
\begin{tabular}{lccl}
\toprule
\textbf{Source} & $k$ & \textbf{Cap} & \textbf{Purpose} \\
\midrule
\citet{altman2023realistic}     & $2$         & $100$         & GNN training \\
\citet{pirmorad2025iclaml}      & ---$^{\dagger}$ & ---     & LLM ICL \\
\textbf{Ours}                   & $\mathbf{2}$ & $\mathbf{50}$ & LLM ICL \\
\bottomrule
\end{tabular}
\caption{Subgraph extraction parameters vs.\ prior work.
Cap $=$ neighbors per expanded account per BFS hop.
$^{\dagger}$Pipeline described as ``extract a localized $k$-hop subgraph''
(their Section II-A) with a $2$-hop illustration (their Figure $2$);
neither $k$ nor a per-hop neighbor cap is stated explicitly.}
\label{tab:subgraph_params}
\end{table}

\paragraph{Context budget.}
We set $k{=}2$ and select at most $50$ unvisited neighboring
accounts per expanded account per hop, pooling incoming and outgoing
counterparties before sampling. The extracted graph retains all
transaction edges among visited accounts. This budget provides
multi-hop context while limiting expansion at high-degree accounts.
Depth and cap are fixed choices, without a cap ablation.
Few-shot demonstrations use $k{=}1$ and at most $15$ neighbors
per account per hop, keeping the $12$-example block to approximately $30$K
tokens. They illustrate the answer format; each test prompt supplies
the subgraph extracted for its assigned transaction.

\subsection{Target Identification in the Original Prompts}
\label{app:prompt_provenance}

The extraction starts from the target's source account, caps
neighbors, and does not force the target edge into the resulting
graph. Matching all $6{,}021$ cases to the original transaction
rows confirms that $324/3{,}753$ HI and $297/2{,}268$ LI graph
strings omit the target transaction. The omitted targets include
$128$ illicit and $196$ benign edges on \hiSmall, and $181$ illicit
and $116$ benign edges on \liSmall. The other
cases contain its transaction attributes but no explicit target
marker: edge IDs are omitted and the task retains a literal
\texttt{<ID>}. Timestamp sorting supplies no target-first convention.
Thus the assigned target is not unambiguously specified to the
model even when its transaction is present.

The text field labeled ``Time span'' counts distinct timestamp
strings rather than elapsed time. The main LLM scores and trace
audit describe responses to these original inputs.

We also group the archived predictions by target presence. Among
target-present benign cases ($2{,}306$ on \hiSmall; $1{,}396$ on
\liSmall), five-run mean flag rates across the seven few-shot
models range from $61.4\%$ to $97.0\%$ and from $77.1\%$ to $98.5\%$,
respectively. Thus, false positives also occur frequently when the
target transaction is present. These are unweighted cohort rates
with all requests retained under the original binary scoring rule.
Targets remain unmarked, and the groups differ in graph size and
other input properties; the comparison does not isolate an effect
of target omission. Per-model rates, run variation, and class-specific
denominators are included in the released analysis files.

\subsection{Target-Marked Inputs and Integrity Checks}
\label{app:target_integrity}

The \texttt{targeted\_prompts\_v1} release is separate from the
original inputs and predictions. It preserves all $6{,}021$ case
IDs, target IDs, subset positions, and weights. Every intended
target matches its original CSV row.

Each new input identifies the target in its instructions and
marks exactly one graph edge. If the target was absent, we insert
it and remove one non-target edge deterministically to preserve
the original edge-count budget. This restores $324$ targets on
\hiSmall and $297$ on \liSmall. Bank-qualified endpoints and
paid/received target fields come from the source records; start,
end, and elapsed time describe the included transactions without
adding a timezone.

Integrity checks verify one marked target, its endpoints and source
fields, unchanged edge count, correct time summary, and absence of
outcome fields and unresolved target placeholders. The release includes
the deterministic builder and validation reports. Scoring labels
remain in the original table.

The version changes target marking, some context-edge membership,
endpoint names, target fields, time summaries, and task instructions;
future comparisons would measure their combined effect.
The source graph and partitions remain unchanged. This version's
effect on predictions and compatibility with model context limits
have not been evaluated.

\section{Model Settings and Compute Budget}
\label{app:models}

Table~\ref{tab:models} reports the model configurations and decoding
parameters used in~\Cref{sec:setup}.

\begin{table*}[!ht]
\centering
\small
\setlength{\tabcolsep}{5pt}
\begin{tabular}{llcccccc}
\toprule
\textbf{Model} & \textbf{Arch} & \textbf{Active} & \textbf{Context} & \textbf{TP} & \textbf{temp} & \textbf{top\_p} & \textbf{top\_k} \\
\midrule
GPT-OSS-20B & MoE & 3.6B & 128K & 1 & 1.0 & 1.0 & --- \\
Nemotron-3-Nano-30B & MoE & 3B & 256K & 2 & 1.0 & 0.95 & --- \\
Qwen3.5-35B-A3B & MoE & 3B & 256K & 2 & 0.6 & 0.95 & 20 \\
GPT-OSS-120B & MoE & 5.1B & 128K & 4 & 1.0 & 1.0 & --- \\
Nemotron-3-Super-120B & MoE & 12B & 256K & 4 & 1.0 & 0.95 & --- \\
Qwen3.5-397B-A17B & MoE & 17B & 256K & 8 & 0.6 & 0.95 & 20 \\
Qwen3.5-27B$^\dagger$ & Dense & 27B & 128K & 4 & 1.0 & 0.95 & 20 \\
\bottomrule
\end{tabular}
\caption{Open-weight model configurations and decoding parameters.
Context = deployed \texttt{max\_model\_len}; TP = tensor-parallel size. All models use
\texttt{max\_tokens}$=32{,}768$.
$^\dagger$Qwen3.5-27B has 256K native context but we deployed at 128K
under $4$-way tensor parallelism (VRAM-bound); this cap drives the $38.7\%$ no-response rate
analyzed in~\Cref{sec:llm:traces}.}
\label{tab:models}
\label{tab:sampling_params}
\end{table*}

The archived GCPAL inputs combine the $79$ GFP and raw-attribute
dimensions with five line-graph summaries: mean and minimum log
positional offset, mean sender-role indicators at both endpoints,
and log line-graph degree. Offsets refer to positions among an
account's transactions in CSV order.

\paragraph{Compute budget.}
LLM inference uses vLLM 0.19+ with reasoning-mode parsing on
H100/H200 (80--141\,GB) nodes. Total compute is
$\approx2{,}840$ GPU-hours, including reruns after vLLM/CUDA crashes.
Qwen3.5-397B-A17B uses $\sim1{,}320$ GPU-hours ($\sim166$ wall-clock
hours), Nemotron-3-Super-120B $\sim610$, and Qwen3.5-35B-A3B $\sim380$.
Recorded summaries cover $433{,}512$ attempted requests across
seven models, two datasets, two main prompts, five stochastic runs,
and the run-42 verification calls: $24.91$\,B input and $0.90$\,B
output tokens, excluding crashed jobs without summaries.
Non-LLM baselines (LightGBM+GFP,
XGBoost+GFP, GCPAL+GFP~\citep{lu2024gcpal}) train on a single H100
(80\,GB). The three total $\sim$$3$ H100-hours across all
$5$ seeds. Scorer inputs and training settings appear in
\Cref{app:non_llm}.

A zero-shot compact-set pass uses $98$--$165$M tokens on \hiSmall
and $51$--$102$M on \liSmall (input plus output, run~$42$),
calculated as mean tokens per case times the selected-set size.
Table~\ref{tab:per_pass_tokens} gives per-model counts alongside
the supplementary DT estimates.

\appendixkeepheading{7\baselineskip}
\section{\ours{} Construction and Sampling Evaluation}
\label{app:coreset_group}

\subsection{Coreset Algorithm Pseudocode}
\label{app:algorithm}

Algorithm~\ref{alg:ht_coreset} gives the procedure referenced in
~\Cref{sec:method:coreset}.

\begin{algorithm}[!htbp]
\caption{\ours: Stratified Coreset + HT Weighting}
\label{alg:ht_coreset}
\begin{algorithmic}[1]
\REQUIRE Test set $\mathcal{E}_{\text{test}}$; rater set
$\mathcal{R}=\mathcal{M}\times\mathcal{D}$ (scorers $\times$ seeds)
with per-rater predictions $\{\hat{p}_r\}_{r\in\mathcal{R}}$; ML
threshold $\tau$; hard-neg ratio $\alpha{=}0.3$
\ENSURE Subset $\mathcal{S}$, importance weights $\{w_i\}_{i\in\mathcal{S}}$
\STATE $\mathcal{S}_{\text{ill}} \leftarrow \{e_i : y_i = 1\}$ \COMMENT{Stage 1: retain all illicit}
\STATE $n_{\text{benign}} \leftarrow 2 \times |\mathcal{S}_{\text{ill}}|$ \COMMENT{smallest evaluated benign budget}
\STATE $\bar{p}(e_i) \leftarrow \frac{1}{|\mathcal{R}|} \sum_{r\in\mathcal{R}} \hat{p}_r(e_i)$ \COMMENT{ensemble mean over scorers $\times$ seeds}
\STATE $d(e_i) \leftarrow 1 - 2\,|\bar{p}(e_i) - 0.5|$ \COMMENT{difficulty: $1$ at $\bar p=0.5$}
\STATE $\mathcal{B} \leftarrow \{e_i : y_i = 0\}$ \COMMENT{benign pool}
\STATE $\mathcal{H} \leftarrow \{e_i \in \mathcal{B} : \bar{p}(e_i) \geq \tau/2\}$ \COMMENT{hard negatives}
\STATE $n_{\text{hard}} \leftarrow \min(\lfloor\alpha\, n_{\text{benign}}\rfloor, |\mathcal{H}|)$
\STATE $\mathcal{S}_{\text{hard}} \leftarrow$ uniform sample of $n_{\text{hard}}$ from $\mathcal{H}$ (w/o replacement)
\STATE Partition $\mathcal{B}\setminus\mathcal{H}$ into terciles $\{T_1,T_2,T_3\}$ by $d(\cdot)$ (33/67 percentiles)
\STATE $q \leftarrow \max\!\left(1,\ \lfloor (n_{\text{benign}}{-}n_{\text{hard}})/3 \rfloor\right)$
\STATE $\mathcal{S}_{\text{strat}} \leftarrow \bigcup_{k=1}^{3}$ uniform sample of $\min(q, |T_k|)$ from $T_k$
\STATE $\mathcal{S}_{\text{benign}} \leftarrow \mathcal{S}_{\text{hard}} \cup \mathcal{S}_{\text{strat}}$
\IF{$|\mathcal{S}_{\text{benign}}| < n_{\text{benign}}$}
  \STATE Fill remainder uniformly from $\mathcal{B}\setminus\mathcal{S}_{\text{benign}}$ into $\mathcal{S}_{\text{fill}}$; $\mathcal{S}_{\text{benign}} \leftarrow \mathcal{S}_{\text{benign}}\cup\mathcal{S}_{\text{fill}}$
\ENDIF
\STATE $\mathcal{S} \leftarrow \mathcal{S}_{\text{ill}} \cup \mathcal{S}_{\text{benign}}$
\FOR{each $e_i \in \mathcal{S}$}
    \STATE $w_i \leftarrow N_{\text{stratum}(i)} / n_{\text{stratum}(i)}$ \COMMENT{$w_i{=}1$ for $e_i\in\mathcal{S}_{\text{ill}}$}
\ENDFOR
\RETURN $\mathcal{S},\ \{w_i\}$
\end{algorithmic}
\end{algorithm}

\paragraph{Budget selection.}
The $12$-point benign-budget grid increases geometrically from
$\max(500,2n_{\mathrm{illicit}})$ to the full benign population.
Both datasets pass the criterion in \Cref{sec:method:coreset} at
the lower bound, giving the released $1{:}2$ ratio. Smaller budgets
were not evaluated. Figure~\ref{fig:reduction_sweep} reports agreement
across the sweep; \Cref{app:estimation} explains its exactness for
the construction ensemble.

\paragraph{Operating thresholds.}
The size sweep uses ensemble thresholds $0.80$ on \hiSmall and
$0.48$ on \liSmall. The archived equal-budget ablation uses a
finer threshold grid, selecting $0.7880904523$ on \hiSmall and
$0.4876884422$ on \liSmall.
Its full-split references, in $(P,R,F1)$ order and expressed as
percentages, are $(93.1079,57.2342,70.8911)$ on \hiSmall and
$(74.0741,18.5185,29.6296)$ on \liSmall. Each deviation in
Table~\ref{tab:ablation} uses that experiment's fixed threshold.
These experiments have different precision references. All operating
thresholds were selected on the full test split.

\subsection{Estimation Properties and Scope}
\label{app:estimation}

We distinguish unbiased count estimation, ratio estimation, and
exact reproduction of the construction ensemble's metrics
(\hyperref[sec:method:estimation]{estimation scope}).

\paragraph{Counts.} Let $\pi_i>0$ be the known inclusion probability
of edge $e_i$ and $w_i=1/\pi_i$. For fixed per-edge predictions,
each full-split confusion count is a total $T=\sum_i z_i$, where
$z_i$ indicates membership in TP, FP, or FN. Its HT estimate is
$\widehat T=\sum_{i\in\mathcal S}z_i/\pi_i$, and
$\mathbb{E}[\widehat T]=\sum_i\pi_i z_i/\pi_i=T$
\citep{horvitz1952generalization}. This expectation is over the
sampling design. Predictions and their input context must be held
fixed across draws; the result does not cover changes in predictions
caused by altering the graph presented to a model.

\paragraph{Ratios.} Precision and F1 are ratios of weighted
counts and need not be unbiased in finite samples. Their sampling
variance also depends on the predictor's false positives in the
sampled strata. Consistency requires concentration of the estimated
counts and a denominator bounded away from zero in the relevant
asymptotic regime~\citep{sarndal1992model}. Recall is exact for
fixed predictions under Stage~1: every illicit edge is retained at
weight one, so both TP and FN are observed in full.

\paragraph{Exact reproduction in the reported experiments.}
The evaluated ensemble flags no down-sampled benign edge at its
operating threshold. Every edge contributing to its TP, FP, or FN
therefore has weight one. The subset and full-split computations
use the same contributing edges: $|\Delta P|=|\Delta R|=0$ and
$|\Delta F1|<10^{-13}$ pp at all $12$ tested subset sizes
(Figure~\ref{fig:reduction_sweep}). This observation depends on the
predictor and the reported sampling configurations. Stratifying on
ensemble difficulty alone does not establish it, as the B4 and B5
results show (Table~\ref{tab:ablation}).

\paragraph{Conditional sampling design.}
For LLM uncertainty, we hold the construction scores, selected
budget, per-edge contexts, and five prediction vectors fixed.
All illicit edges and hard negatives are censused at the released
budget. Only the three easy-benign terciles are sampled. Their
population sizes are $(334{,}638,344{,}779,334{,}638)$ on
\hiSmall and $(456{,}649,470{,}487,456{,}649)$ on \liSmall;
realized sample sizes are $(713,713,713)$ and $(415,414,414)$.
On \liSmall, the residual budget adds one uniformly sampled edge
after the initial allocation. Conditional on the final stratum
counts, each within-stratum sample is uniform without replacement,
and the three samples are independent. Equal weights in two
terciles do not make them one stratum.

\paragraph{Design-based standard errors.}
Let $z_{ik}$ indicate an illicit prediction on benign edge $i$
for inference run $k$. For a sampled stratum $h$, write $N_h$
and $n_h$ for its population and sample sizes, and $s_{hk}^2$
for its sample variance (denominator $n_h-1$, with $n_h\geq2$).
Census strata contribute zero variance. Then
\begin{align}
 \widehat{\mathrm{FP}}_k
 &=\mathrm{FP}_{\mathrm{hard},k}+\sum_h N_h\bar z_{hk},\\
 \widehat{\mathrm{Var}}(\widehat{\mathrm{FP}}_k)
 &=\sum_h \frac{N_h^2}{n_h}
       \left(1-\frac{n_h}{N_h}\right)s_{hk}^2.
\end{align}
TP and FN are observed in full. Precision and F1 are functions
$g_k$ of the estimated FP total. Where their denominators are
nonzero, their derivatives are
\begin{align}
 P'_k&=-\frac{\mathrm{TP}_k}
                   {(\mathrm{TP}_k+\widehat{\mathrm{FP}}_k)^2},\\
 F1'_k&=-\frac{2\mathrm{TP}_k}
       {(2\mathrm{TP}_k+\mathrm{FN}_k+\widehat{\mathrm{FP}}_k)^2}.
\end{align}
We retain the reported mean of five metrics, rather than pooling
predictions before scoring. Its linearized edge value is
$u_i=\frac{1}{5}\sum_{k=1}^{5}g'_k z_{ik}$; applying the same
stratified variance formula to $u_i$ gives an approximate sampling
SE~\citep{sarndal1992model}. This includes cross-run covariance
because the runs share the sampled edges. For FS--ZS differences,
we apply the formula to $u_i^{\mathrm{FS}}-u_i^{\mathrm{ZS}}$,
retaining same-edge covariance across prompts as well.
Lift scales precision by the known full-test prevalence.
Recall and illicit-only typology F1 have zero sampling variance
under these conditions. Run SD, reported separately with
denominator $5-1$, describes inference variation on the observed
coreset; it is not combined with the sampling SE.

\paragraph{Finite-population confidence intervals.}
Some sampled strata contain only positive predictions. Their
plug-in variance is zero even though predictions on unobserved
edges need not be constant. We therefore use a separate,
conservative interval construction. If $x_{hk}$ of $n_h$ sampled
edges are flagged, then
$X_{hk}\sim\mathrm{Hypergeom}(N_h,M_{hk},n_h)$, where
$M_{hk}$ is the unknown full-stratum flagged count. We invert
equal-tailed tests, retaining integer counts $M_{hk}$ for which
both $\Pr(X_{hk}\geq x_{hk})$ and
$\Pr(X_{hk}\leq x_{hk})$ are at least $\alpha_c/2$.
For each five-run cell, $\alpha_c=0.05/15$ allocates error
across three strata and five runs. Summing stratum endpoints
with the known hard-negative FP count bounds each run's FP
total. Since precision and F1 decrease with FP, transforming
the endpoints reverses their order; averaging the five lower
and upper endpoints bounds the mean metric.

The union bound gives at least $95\%$ coverage for each cell,
including boundary samples, without assuming independent runs.
For an FS--ZS contrast, we instead allocate $0.05/30$ across
both prompts and subtract opposite metric endpoints. This
conservative contrast interval does not exploit pairing, unlike
the linearized SE. Coverage is per cell or contrast, not
simultaneous across the complete results tables. It concerns the
five fixed prediction vectors under the frozen sampling design,
not future inference runs or new populations. This calculation
uses cached predictions; it is not an empirical comparison
against full-test LLM outputs or independent coreset draws.

\subsection{\naive vs \ours}
\label{app:naive_vs_ht}

\naive uses approximately $6$M evaluations on \hiSmall and
$1{,}000$ Optuna trials on \liSmall. The search retains all eight
typologies and limits KL divergence from the full split's illicit-edge
distributions to $0.20$ for typology and $0.15$ for difficulty.
It requires $|\Delta F1|\leq5$ pp, without separate precision or
recall constraints. Draws must also pass our item response theory
(IRT) eligibility screen: at least $100$ illicit edges, at least
five model--seed raters, illicit-edge difficulty SD of at least
$0.10$, and at least five illicit edges in each of three percentile
intervals of difficulty. The smallest feasible draws found
contain $641{,}389$ edges on \hiSmall and $1{,}141{,}186$ on
\liSmall, with F1 deviations of $5.0$ and $4.9$ pp. These are
$1.6\times$ and $1.2\times$ reductions in target count.

Preserving the full split's class ratio leaves only about
$3{,}753\times0.0012\approx4$ illicit edges at the released
\hiSmall budget. Even the selected $641$K-edge draw retains just
$790/1{,}251\approx63\%$ of illicit edges, so recall varies with the
sample. HT-Coreset avoids this source of variation by retaining
all illicit edges.

\subsection{Construction-Ensemble Comparison}
\label{app:construction_comparison}

Table~\ref{tab:ablation} compares seven alternative samplers with
\ours{} at the released subset sizes. This experiment evaluates
the original construction ensemble; the family-held-out study in
\Cref{sec:method:ablation} evaluates predictors excluded from
stratum construction. The full budget sweep appears in
Figure~\ref{fig:reduction_sweep}.

B1 (uniform) and B2 (class-stratified) test ordinary subsampling.
B3 retains all illicit edges and samples benign edges uniformly
without weights. B4 uses weighted difficulty strata without
hard-negative oversampling. B5 uses Neyman allocation~\citep{neyman1934allocation} over score
strata~\citep{fogliato2024reducedlabel}; B6 uses account-graph
random walks~\citep{leskovec2006sampling}; B7 adapts stratified
evaluation~\citep{gao2019efficient} to payment-format groups.
The last three test allocation by model scores, graph topology,
and transaction attributes, respectively. Each comparison uses
$50$ draws with the original construction ensemble.

\begin{table}[t]
\centering
\small
\setlength{\tabcolsep}{3pt}
\begin{tabular}{lcccr}
\toprule
& $|\Delta$F1$|$ & $|\Delta$P$|$ & $|\Delta$R$|$ & $\sigma$ \\
\midrule
\multicolumn{5}{l}{\textit{\hiSmall (Full: P=93.1, R=57.2, F1=70.9)}} \\
B1 Random       & 18.1 & 15.6 & 22.3 & 24.7 \\
B2 Stratified   & 16.0 & 11.1 & 18.0 & 19.9 \\
B3 No IW        &  1.9 &  6.9 &  0.0 &  0.0 \\
B4 No Hard-Neg  &  2.6 &  8.7 &  0.0 &  3.3 \\
B5 Fogliato     &  6.3 &  3.7 &  8.7 &  7.9 \\
B6 Leskovec     & 23.7 & 22.2 & 24.5 & 30.0 \\
B7 Gao          & 20.6 & 14.7 & 24.0 & 26.1 \\
\textbf{\ours} & \textbf{0.0} & \textbf{0.0} & \textbf{0.0} & \textbf{0.0} \\
\midrule
\multicolumn{5}{l}{\textit{\liSmall (Full: P=74.1, R=18.5, F1=29.6)}} \\
B1 Random       & 34.9 & 63.4 & 28.4 & 36.1 \\
B2 Stratified   & 42.8 & 57.7 & 39.9 & 46.7 \\
B3 No IW        &  1.6 & 25.9 &  0.0 &  0.0 \\
B4 No Hard-Neg  &  1.6 & 25.9 &  0.0 &  0.0 \\
B5 Fogliato     & 11.2 &  9.8 & 10.8 & 14.8 \\
B6 Leskovec     & 32.1 & 67.3 & 23.2 & 28.3 \\
B7 Gao          & 33.1 & 66.4 & 23.9 & 30.3 \\
\textbf{\ours} & \textbf{0.0} & \textbf{0.0} & \textbf{0.0} & \textbf{0.0} \\
\bottomrule
\end{tabular}
\caption{Construction ablation for the original three-scorer ensemble
($k{=}50$ draws per method). On \hiSmall, $n{=}3{,}753$ and the
threshold is $0.7881$; on \liSmall, $n{=}2{,}268$ and it is $0.4877$.
All $\Delta$ values and $\sigma$ (std of detection F1) are in pp.
\ours{} retains every edge contributing to this ensemble's
TP, FP, and FN, giving exact P/R and numerical agreement on F1
($|\Delta$F1$|<10^{-13}$ pp).}
\label{tab:ablation}
\end{table}

B1--B2 capture only $1$--$5$ illicit edges, with F1 varying
across draws ($\sigma{>}19$ pp). B3 and B4 retain all illicit
edges but deviate in precision by $6.9$--$25.9$ pp. B3 changes
both the benign sampler and weighting, so it is not a
weight-only ablation. B5 captures only about $49$ of the $1{,}251$
illicit edges on \hiSmall. B6 and B7 ignore class labels and have large
detection-metric deviations. \ours{} has the
smallest deviations in this comparison; per-model results appear
in \Cref{app:permodel}. Retention and draw-distribution plots are
included in the release under \path{results/figures/}, with the
underlying draws in \path{results/coreset/baselines_draws.csv}.

\subsection{Sampling with a Held-Out Predictor Family}
\label{app:heldout_sampling}

This study uses cached full-test probabilities. In each direction, one GFP booster
family constructs the strata and the other supplies the evaluated
predictions. The constructor score is its five-seed probability mean,
and its operating threshold is the mean of its saved
validation-selected thresholds. Each evaluated seed keeps its own
validation-selected threshold. The constructor never uses the
other family's scores or errors.

We fix benign budgets at $2$, $4$, and $8$ times the illicit count,
retain every illicit edge, and use $500$ independent draws for each
configuration. Uniform sampling uses one benign stratum. Difficulty
sampling uses the $33$rd and $67$th percentiles of
$d(e)=1-2|\bar p(e)-0.5|$ from the constructor.
HT-Coreset additionally reserves up to $30\%$ of the benign budget
for scores at least half the constructor threshold. Each nonempty
stratum has positive inclusion probability. To fix allocation across
draws, remaining slots use a deterministic largest-remainder rule
proportional to residual stratum capacities. This replaces the
released draw's random residual fill; its original indices and
weights are unchanged.

Each draw is shared across the five evaluated seeds. We compare
the mean of their weighted metrics with the mean of their exact
full-test metrics. Signed bias and RMSE summarize errors across
draws, rather than variation across training seeds. Confidence
intervals use the hypergeometric count construction above, with
$0.05/(5H)$ allocated to each count for $H$ noncensus benign
strata. We report coverage and width together: conservative
coverage need not imply a precise estimate. Coverage is for each
fixed five-seed configuration, not simultaneous over the experiment.

Table~\ref{tab:heldout_sampling} reports the released-size budgets;
Table~\ref{tab:heldout_sampling_budgets} adds the larger budgets.
The released CSVs include per-draw and per-seed results, fixed
protocol settings, and input hashes. Both predictor families share
GFP features; this experiment does not evaluate full-test LLM
predictions or transfer to arbitrary predictors.

\begin{table*}[!ht]
\centering\small
\setlength{\tabcolsep}{5pt}
\begin{tabular}{lrrrrr}
\toprule
\textbf{Sampler} & \textbf{P RMSE} & \textbf{F1 bias} & \textbf{F1 RMSE} & \textbf{F1 width} & \textbf{Cov.} \\
\midrule
\multicolumn{6}{l}{\textit{HI-Small: LGB $\to$ XGB, benign budget $4\times$ illicit count}} \\
Uniform + HT & 14.46 & +0.48 & 5.04 & 29.6 & 100.0 \\
Terciles + HT & 13.84 & +0.63 & 4.74 & 49.6 & 100.0 \\
HT-Coreset & 0.00 & +0.00 & 0.00 & 44.9 & 100.0 \\
\addlinespace[2pt]
\multicolumn{6}{l}{\textit{HI-Small: LGB $\to$ XGB, benign budget $8\times$ illicit count}} \\
Uniform + HT & 10.63 & +0.38 & 3.64 & 21.0 & 100.0 \\
Terciles + HT & 10.75 & +0.43 & 3.63 & 38.6 & 99.8 \\
HT-Coreset & 0.00 & +0.00 & 0.00 & 32.9 & 100.0 \\
\addlinespace[2pt]
\multicolumn{6}{l}{\textit{HI-Small: XGB $\to$ LGB, benign budget $4\times$ illicit count}} \\
Uniform + HT & 15.28 & +0.48 & 5.40 & 30.2 & 100.0 \\
Terciles + HT & 14.99 & +0.42 & 5.32 & 49.4 & 100.0 \\
HT-Coreset & 3.59 & -0.05 & 1.44 & 43.6 & 100.0 \\
\addlinespace[2pt]
\multicolumn{6}{l}{\textit{HI-Small: XGB $\to$ LGB, benign budget $8\times$ illicit count}} \\
Uniform + HT & 10.90 & +0.21 & 3.80 & 22.0 & 100.0 \\
Terciles + HT & 11.16 & +0.51 & 3.86 & 38.6 & 100.0 \\
HT-Coreset & 2.60 & -0.02 & 0.99 & 32.1 & 100.0 \\
\addlinespace[2pt]
\multicolumn{6}{l}{\textit{LI-Small: LGB $\to$ XGB, benign budget $4\times$ illicit count}} \\
Uniform + HT & 36.12 & +0.71 & 4.07 & 23.1 & 100.0 \\
Terciles + HT & 36.13 & +0.91 & 3.94 & 28.4 & 100.0 \\
HT-Coreset & 2.21 & +0.04 & 0.50 & 26.1 & 99.6 \\
\addlinespace[2pt]
\multicolumn{6}{l}{\textit{LI-Small: LGB $\to$ XGB, benign budget $8\times$ illicit count}} \\
Uniform + HT & 32.18 & +0.38 & 3.25 & 18.7 & 99.6 \\
Terciles + HT & 32.52 & +0.52 & 3.23 & 26.0 & 100.0 \\
HT-Coreset & 2.24 & +0.04 & 0.38 & 23.7 & 100.0 \\
\addlinespace[2pt]
\multicolumn{6}{l}{\textit{LI-Small: XGB $\to$ LGB, benign budget $4\times$ illicit count}} \\
Uniform + HT & 43.19 & +1.58 & 4.93 & 23.7 & 99.6 \\
Terciles + HT & 41.42 & +0.76 & 5.09 & 29.0 & 100.0 \\
HT-Coreset & 10.20 & +0.37 & 2.22 & 25.9 & 100.0 \\
\addlinespace[2pt]
\multicolumn{6}{l}{\textit{LI-Small: XGB $\to$ LGB, benign budget $8\times$ illicit count}} \\
Uniform + HT & 35.51 & +0.67 & 4.03 & 19.6 & 99.6 \\
Terciles + HT & 36.66 & +0.95 & 4.01 & 26.7 & 100.0 \\
HT-Coreset & 9.69 & +0.20 & 1.84 & 23.6 & 100.0 \\
\addlinespace[2pt]
\bottomrule\end{tabular}
\caption{Larger budgets in the held-out-family study ($500$ draws per configuration).
Bias is mean estimated F1 minus exact full-test F1. RMSE, bias, and mean $95\%$ interval
width are in percentage points; coverage is a percentage. Every quantity concerns the
mean of the five fixed seed metrics. The primary $1{:}2$ budgets appear in
Table~\ref{tab:heldout_sampling}; all budgets and per-seed results are released as CSVs.}
\label{tab:heldout_sampling_budgets}
\end{table*}

\subsection{Supervised Baselines and Data Splits}
\label{app:non_llm}

\paragraph{Scorer inputs and training.}
LightGBM+GFP, XGBoost+GFP, and GCPAL+GFP all use $73$ GFP
features~\citep{blanusa2024gfp} and $6$ raw edge attributes.
GCPAL adds $5$ line-graph features, giving $84$ inputs in total.
Our GCPAL reproduction~\citep{lu2024gcpal} keeps its GIN
encoder~\citep{xu2019powerful} and contrastive pre-training.
We adapt its transaction-node classification setup to AMLworld
by constructing a line graph with GFP features.
The boosters train on the training partition and are Optuna-tuned
over $50$ trials with a fixed search grid (learning rate
$\in[10^{-3}, 10^{-1}]$, depth/leaves in standard ranges).
GCPAL uses the architecture and schedule reported by its authors,
with no search of our own. Its different fine-tuning split is
documented below. \Cref{app:models} reports compute costs.

Table~\ref{tab:non_llm} reports the supervised baselines referenced in
~\Cref{sec:llm:results}. The rows report full-test detection metrics, mean$\pm$std
across five seeds. Individual boosters use their validation-selected
thresholds; the two-booster ensemble uses the inherited fixed thresholds
$0.80$ on \hiSmall and $0.48$ on \liSmall. The primary task reference averages probabilities
across both boosters and all five seeds before thresholding,
giving pooled $F1$ of $68.1\%$ on \hiSmall and $28.6\%$ on \liSmall.

\begin{table}[!htbp]
\centering
\small
\setlength{\tabcolsep}{3pt}
\begin{tabular}{lccc}
\toprule
\textbf{Method} & \textbf{F1} & \textbf{P} & \textbf{R} \\
\midrule
\multicolumn{4}{c}{\textit{\hiSmall}} \\
\midrule
LightGBM+GFP    & 65.7\tiny{$\pm$1.2} & 77.9\tiny{$\pm$4.4} & 56.9\tiny{$\pm$0.7} \\
XGBoost+GFP     & 67.7\tiny{$\pm$0.3} & 81.7\tiny{$\pm$2.1} & 57.8\tiny{$\pm$0.6} \\
Two boosters & \textbf{67.8}\tiny{$\pm$0.2} & \textbf{84.3}\tiny{$\pm$0.5} & 56.7\tiny{$\pm$0.2} \\
\midrule
\multicolumn{4}{c}{\textit{\liSmall}} \\
\midrule
LightGBM+GFP    & 27.5\tiny{$\pm$0.5} & 50.0\tiny{$\pm$4.9} & 19.0\tiny{$\pm$0.4} \\
XGBoost+GFP     & 28.6\tiny{$\pm$0.3} & 63.1\tiny{$\pm$5.0} & 18.5\tiny{$\pm$0.3} \\
Two boosters & \textbf{28.6}\tiny{$\pm$0.2} & 59.8\tiny{$\pm$1.2} & 18.8\tiny{$\pm$0.1} \\
\bottomrule
\end{tabular}
\caption{Supervised detection baselines on the full file-order
test partition (mean$\pm$std, five training seeds). Means are percentages;
stds are in pp.}
\label{tab:non_llm}
\end{table}

\paragraph{Split provenance.}
The archived test labels match the final $20\%$ of rows in each
released CSV. The code does not sort transactions by timestamp
before partitioning. Train, validation, and test timestamp ranges
overlap. Sorting first would preserve $99.73\%$ of test membership
on \hiSmall and $99.84\%$ on \liSmall, but change illicit counts
from $1{,}251$ to $1{,}797$ on \hiSmall and from $756$ to $925$ on
\liSmall. Thus the small overall membership change does not
establish negligible impact. We have not retrained on strictly
chronological partitions.

\paragraph{Fixed-threshold removal check.}
At the inherited thresholds ($0.80$ on \hiSmall and $0.48$ on
\liSmall), removing GCPAL from the
pooled three-scorer prediction changes detection $F1$ from
$70.87\%$ to $68.07\%$ on \hiSmall and $29.57\%$ to $28.63\%$ on \liSmall.
These thresholds were selected on the full test set for the original
construction scorer and are held fixed, without retuning.
Our archived GCPAL reproduction~\citep{lu2024gcpal} used a random
fine-tuning split, which overlaps about $60\%$ of the
test set. Its scores remain part of the frozen construction,
but GCPAL is excluded from the primary task comparisons.

The two-booster reference also has exact full-test and HT-weighted
confusion counts at these thresholds. Writing its score as $p_2$
and the construction score as $p_3$, nonnegative GCPAL probabilities
give $p_3\geq 2p_2/3$. Thus any benign edge with $p_2\geq\tau$
also has $p_3>\tau/2$ and belongs to the original hard-negative
census. All illicit edges were already retained. This argument is
specific to the pooled scores and inherited thresholds; it does
not redefine the original strata or extend to arbitrary predictors.

\subsection{Per-Model Sampling Comparison}
\label{app:permodel}

Table~\ref{tab:permodel_ablation} reports mean $|\Delta F1|$
for each sampling method and constituent model over five training
seeds on one canonical draw per method. The exact construction-ensemble result in the construction ablation
does not extend to every constituent scorer. HT-Coreset has
small deviations for LightGBM and XGBoost, including three values
of $0.007$, $0.012$, and $0.045$ pp that round to $0.0$ in the
table. GCPAL deviations are $3.9$ pp on \hiSmall and $2.6$ pp on \liSmall.
B3 samples benign edges uniformly without weights; B4 uses weighted
difficulty strata without hard-negative oversampling. Both retain
all illicit edges, so recall is exact for fixed predictions.

\begin{table}[!htbp]
\centering
\small
\setlength{\tabcolsep}{3pt}
\begin{tabular}{lccc}
\toprule
\textbf{Method} & \textbf{LightGBM} & \textbf{XGBoost} & \textbf{GCPAL} \\
\midrule
\multicolumn{4}{c}{\textit{\hiSmall}} \\
\midrule
B1 Random       &  1.1 &  2.5 &  7.5 \\
B2 Stratified   & 15.8 & 19.8 &  9.1 \\
B3 No IW        &  3.0 &  2.3 &  5.3 \\
B4 No Hard-Neg  &  4.5 &  2.3 &  5.3 \\
B5 Fogliato     &  7.7 &  8.2 &  1.4 \\
B6 Leskovec     & 67.6 & 69.1 & 59.2 \\
B7 Gao          &  5.1 &  2.5 &  7.4 \\
\textbf{\ours} & \textbf{0.0} & \textbf{0.0} & 3.9 \\
\midrule
\multicolumn{4}{c}{\textit{\liSmall}} \\
\midrule
B1 Random       & 28.2 & 28.9 & 17.5 \\
B2 Stratified   & 71.8 & 71.1 & 63.8 \\
B3 No IW        &  3.1 &  1.9 &  4.6 \\
B4 No Hard-Neg  &  3.1 &  1.9 &  4.0 \\
B5 Fogliato     &  6.4 & 10.2 & 12.8 \\
B6 Leskovec     & 11.8 & 11.1 & 17.5 \\
B7 Gao          & 28.2 & 28.9 & 17.5 \\
\textbf{\ours} & \textbf{0.0} & \textbf{0.0} & 2.6 \\
\bottomrule
\end{tabular}
\caption{Per-model $|\Delta F1|$ (pp), averaged over five training
seeds on one canonical draw per sampling method. Each cell compares with that individual scorer's
full-set F1; Table~\ref{tab:ablation} instead uses the construction
ensemble. All models use GFP features. HT-Coreset's rounded zeros
include small nonzero deviations; only its XGBoost result on
\hiSmall matches to numerical precision ($4{\times}10^{-15}$ pp).}
\label{tab:permodel_ablation}
\end{table}

\subsection{Sampling Evaluation on Elliptic}
\label{app:elliptic}

We also evaluate metric agreement outside synthetic AMLworld by
running \ours and the seven ablation baselines on the
Elliptic Bitcoin Dataset~\citep{weber2019anti}: $203{,}769$ transactions over $49$ time
steps, $234{,}355$ directed UTXO edges, $4{,}545$ illicit and $42{,}019$
licit transactions in the labeled subset (the remaining ${\sim}77\%$ are
unknown and reserved for $k$-hop context). We follow the original
temporal split (train: time-steps 1--34; test: 35--49) and use the
$165$ transaction features released with the dataset, so no feature
engineering is shared with the AMLworld pipeline.

\paragraph{Setup.}
We train a two-model soft-vote ensemble using LightGBM and XGBoost,
the same two model types as the primary AMLworld evaluation reference, calibrate the threshold on a 10\% held-out validation
slice of the train split, and apply \ours{} with the same hyperparameters
($2{:}1$ benign:illicit budget, up to $30\%$ hard-negative allocation, three
difficulty terciles). Each of the seven matched-size ablations (B1--B7)
uses $k{=}50$ draws. Code:
\texttt{scripts/elliptic/} (download $\to$ prepare $\to$ ML $\to$ \ours{} $\to$
eval); a single SLURM script (\path{scripts/slurm/20_elliptic_pipeline.sh})
reproduces every number in $\sim$1--2 hours on a 16-core CPU node.

\paragraph{Results.}
On the labeled test split (16{,}670 nodes, 1{,}083 illicit, illicit rate
$6.50\%$), the LightGBM\,$+$\,XGBoost ensemble reaches $\text{P}=89.99\%$,
$\text{R}=73.04\%$, $F1=80.63\%$ at threshold $0.614$ using
probabilities pooled across five seeds; the seed-level mean is
$F1=80.39\%$ (SD $0.79$ pp). \ours retains all $1{,}083$ illicit
nodes plus $2{,}166$ benign ($n{=}3{,}249$, a $5.1\times$ reduction over
the full labeled test) and yields
$|\Delta\text{P}|{=}|\Delta\text{R}|{=}0.00$ pp exactly and
$|\Delta F1|{<}10^{-13}$ pp.
Table~\ref{tab:elliptic_ablation} reports the seven matched-size
ablations. Six have nonzero mean absolute deviations on at least
one metric (Wilcoxon $p<10^{-4}$; paired-bootstrap $95\%$ CI for
$|\Delta F1|$ excludes zero). B3 also reproduces the full-split
metrics here: it applies unit weights to nodes selected by the
\ours{} rule, and the evaluated ensemble's false positives all lie
in the fully retained hard-negative stratum. AMLworld B3 instead
draws benign edges uniformly, so the B3 rows do not isolate a
dataset effect. B4 allocates no sample
to that stratum. It therefore omits those false positives in every
draw, producing a fixed deviation rather than sampling variation;
its hard-negative nodes have zero inclusion probability.

\begin{table}[!htbp]
\centering
\small
\setlength{\tabcolsep}{3pt}
\begin{tabular}{lccccr}
\toprule
& $|\Delta F1|$ & $|\Delta$P$|$ & $|\Delta$R$|$ & $\sigma$ & $p_W$ \\
\midrule
B1 Random       & $\phantom{0}1.57$ & $\phantom{0}1.42$ & $\phantom{0}2.22$ & $1.85$ & $<10^{-4}$ \\
B2 Stratified   & $\phantom{0}1.62$ & $\phantom{0}1.77$ & $\phantom{0}2.13$ & $2.02$ & $<10^{-4}$ \\
B3 No IW        & $\phantom{0}0.00$ & $\phantom{0}0.00$ & $\phantom{0}0.00$ & $0.00$ & --- \\
B4 No Hard-Neg  & $\phantom{0}3.79$ & $10.01$ & $\phantom{0}0.00$ & $0.00$ & $<10^{-4}$ \\
B5 Fogliato     & $\phantom{0}0.93$ & $\phantom{0}0.64$ & $\phantom{0}1.48$ & $1.14$ & $<10^{-4}$ \\
B6 Leskovec     & $\phantom{0}1.01$ & $\phantom{0}2.56$ & $\phantom{0}0.00$ & $0.90$ & $<10^{-4}$ \\
B7 Gao          & $\phantom{0}0.91$ & $\phantom{0}2.28$ & $\phantom{0}0.00$ & $1.12$ & $<10^{-4}$ \\
\textbf{\ours} & \textbf{$\phantom{0}0.00$} & \textbf{$\phantom{0}0.00$} & \textbf{$\phantom{0}0.00$} & \textbf{$0.00$} & --- \\
\bottomrule
\end{tabular}
\caption{Elliptic real-data ablation
($n{=}3{,}249$, $k{=}50$ draws). All $\Delta$ and $\sigma$ are in pp;
$\sigma$ = std of subset $F1$; $p_W$ = one-sided Wilcoxon vs \ours{} ($p$ undefined for B3
because every draw has the same metrics as \ours{} on this dataset).
B3 follows the Elliptic HT selection rule with unit weights; its
AMLworld counterpart instead samples benign edges uniformly.
B4 excludes the hard-negative stratum and cannot estimate its
contribution to full-test counts.
The $0.00$ entries in the $|\Delta F1|$ column (\ours{} and B3) are
likewise $<10^{-13}$ pp; the $|\Delta$P$|$ and $|\Delta$R$|$ zeros are
exact.}
\label{tab:elliptic_ablation}
\end{table}

\paragraph{Scope of the external check.}
Elliptic tests the construction with transaction nodes, native
features, and public Bitcoin data. For fixed predictions, HT count
estimation requires known, positive inclusion probabilities.
The evaluated ensemble's exact agreement follows from retaining all
nodes contributing to its confusion counts.
Elliptic B6/B7 also retain all illicit nodes, unlike their AMLworld
adaptations. B6 uses an illicit-seeded random walk and a common
benign weight, not known inverse inclusion probabilities; the HT
guarantee does not apply to that comparator.

\section{Full LLM Results}
\label{app:llm_results_group}

\begin{table*}[!ht]
\centering
\small
\setlength{\tabcolsep}{4pt}
\begin{tabular}{llrrrrrrrr}
\toprule
& & \multicolumn{3}{c}{\shortstack{\textbf{HT-weighted}\\(full test)}} & \multicolumn{2}{c}{\shortstack{\textbf{Unweighted}\\(subset)}} & \multicolumn{3}{c}{Not reweighted} \\
\cmidrule(lr){3-5}\cmidrule(lr){6-7}\cmidrule(lr){8-10}
\textbf{Model} & \textbf{M} & $F1_{\text{HT}}$ & $P_{\text{HT}}$ & \textbf{Lift} & F1 $\pm$ SD & P & \textbf{R} & \textbf{TF1} & \textbf{V\%} \\
\midrule
\multicolumn{10}{c}{\textit{\hiSmall}: population illicit rate $0.1232\%$, predict-all floor $F1_{\text{HT}}{=}0.2460\%$} \\
\midrule
GPT-OSS-20B & FS & 0.2360 & 0.1182 & 0.959 & $46.4\pm0.40$ & 32.8 & 79.2 & 10.5 & 88.3 \\
GPT-OSS-20B & ZS & 0.2457 & 0.1230 & 0.999 & $49.9\pm0.08$ & 33.3 & 99.5 & 5.6 & 99.8 \\
Qwen3.5-27B & FS & 0.2404 & 0.1205 & 0.978 & $40.7\pm0.08$ & 32.5 & 54.5 & 10.1 & 61.5 \\
Qwen3.5-27B & ZS & 0.2302 & 0.1153 & 0.936 & $45.8\pm0.05$ & 31.6 & 82.9 & 6.3 & 88.1 \\
Nem-Nano-30B & FS & 0.2454 & 0.1229 & 0.998 & $49.5\pm0.19$ & 33.4 & 95.4 & 7.5 & 98.0 \\
Nem-Nano-30B & ZS & 0.2460 & 0.1231 & 1.000 & $50.0\pm0.02$ & 33.3 & 99.9 & 6.6 & 100.0 \\
Qwen3.5-35B-A3B & FS & 0.2525 & 0.1265 & 1.027 & $48.5\pm0.30$ & 34.0 & 84.9 & 16.5 & 99.1 \\
Qwen3.5-35B-A3B & ZS & 0.2463 & 0.1233 & 1.001 & $49.7\pm0.19$ & 33.3 & 97.8 & 6.1 & 100.0 \\
GPT-OSS-120B & FS & 0.2494 & 0.1249 & 1.014 & $50.3\pm1.60$ & 34.0 & 96.4 & 10.1 & 97.9 \\
GPT-OSS-120B & ZS & 0.2457 & 0.1230 & 0.999 & $49.9\pm0.00$ & 33.3 & 99.8 & 8.1 & 99.8 \\
Nem-Super-120B & FS & 0.2474 & 0.1239 & 1.006 & $47.9\pm0.32$ & 33.7 & 82.4 & 12.3 & 98.0 \\
Nem-Super-120B & ZS & 0.2472 & 0.1237 & 1.005 & $50.0\pm0.10$ & 33.5 & 99.0 & 7.8 & 100.0 \\
Qwen3.5-397B & FS & 0.2507 & 0.1255 & 1.019 & $50.6\pm0.10$ & 34.1 & 98.3 & 18.4 & 99.0 \\
Qwen3.5-397B & ZS & 0.2478 & 0.1241 & 1.007 & $50.2\pm0.03$ & 33.5 & 100.0 & 7.6 & 100.0 \\
Predict-all illicit & --- & 0.2460 & 0.1232 & 1.000 & $50.0\pm0.00$ & 33.3 & 100.0 & --- & --- \\
\midrule
\multicolumn{10}{c}{\textit{\liSmall}: population illicit rate $0.0546\%$, predict-all floor $F1_{\text{HT}}{=}0.1091\%$} \\
\midrule
GPT-OSS-20B & FS & 0.0982 & 0.0491 & 0.900 & $45.5\pm0.28$ & 31.6 & 81.3 & 9.8 & 91.5 \\
GPT-OSS-20B & ZS & 0.1093 & 0.0547 & 1.002 & $50.0\pm0.09$ & 33.4 & 99.7 & 4.1 & 100.0 \\
Qwen3.5-27B & FS & 0.0965 & 0.0483 & 0.884 & $41.0\pm0.17$ & 30.5 & 62.6 & 2.8 & 73.8 \\
Qwen3.5-27B & ZS & 0.0878 & 0.0439 & 0.805 & $40.7\pm0.04$ & 28.4 & 71.8 & 2.5 & 85.0 \\
Nem-Nano-30B & FS & 0.1084 & 0.0542 & 0.993 & $49.6\pm0.18$ & 33.3 & 97.7 & 4.7 & 99.7 \\
Nem-Nano-30B & ZS & 0.1091 & 0.0546 & 1.000 & $50.0\pm0.07$ & 33.3 & 99.8 & 4.5 & 100.0 \\
Qwen3.5-35B-A3B & FS & 0.1067 & 0.0534 & 0.978 & $47.2\pm0.29$ & 32.6 & 85.2 & 10.8 & 100.0 \\
Qwen3.5-35B-A3B & ZS & 0.1088 & 0.0545 & 0.997 & $49.5\pm0.16$ & 33.2 & 96.9 & 3.6 & 100.0 \\
GPT-OSS-120B & FS & 0.1083 & 0.0542 & 0.993 & $49.7\pm2.35$ & 33.6 & 96.0 & 6.6 & 98.3 \\
GPT-OSS-120B & ZS & 0.1092 & 0.0547 & 1.001 & $50.0\pm0.03$ & 33.4 & 100.0 & 6.7 & 100.0 \\
Nem-Super-120B & FS & 0.1089 & 0.0545 & 0.998 & $49.6\pm0.36$ & 34.0 & 91.9 & 9.8 & 99.7 \\
Nem-Super-120B & ZS & 0.1096 & 0.0548 & 1.004 & $50.0\pm0.18$ & 33.4 & 98.7 & 4.6 & 100.0 \\
Qwen3.5-397B & FS & 0.1105 & 0.0553 & 1.012 & $50.6\pm0.13$ & 34.0 & 99.5 & 13.0 & 100.0 \\
Qwen3.5-397B & ZS & 0.1100 & 0.0550 & 1.008 & $50.2\pm0.06$ & 33.5 & 99.9 & 4.3 & 100.0 \\
Predict-all illicit & --- & 0.1091 & 0.0546 & 1.000 & $50.0\pm0.00$ & 33.3 & 100.0 & --- & --- \\
\bottomrule
\end{tabular}
\caption{Full LLM results on both test partitions, means over
five stochastic runs. Both summaries use the same HT-Coreset targets
and predictions. Metrics are percentages except Lift
($P_{\rm HT}$ divided by full-test prevalence). Unweighted F1 includes
its run SD in percentage points; HT-F1 SD and sampling intervals
appear in Table~\ref{tab:llm_sampling_uncertainty}. R and TF1 are
unchanged by weighting because all illicit targets are retained. V\% is the
fraction of requests with positive recorded token usage, including
some without a final answer.}
\label{tab:llm_ht_full}
\end{table*}

\paragraph{Typology extraction.}
We normalize Unicode hyphens in free-text answers and match
explicit typology names in order, checking specific multi-word
labels first. This recovers labels missed by an ASCII-only
matcher; generic uses of ``layer'' are not mapped to ``stack''.
The same parser supplies typology scores in
Tables~\ref{tab:llm_results} and~\ref{tab:llm_ht_full}.
Binary detection metrics are unaffected by typology extraction.

\paragraph{Scoring convention.}
TF1 uses only illicit edges with known reference typologies
($791$ on \hiSmall and $174$ on \liSmall). A benign verdict maps to
\texttt{none}, regardless of any named typology; missing or
unrecognized typologies also map to \texttt{none}. The archived
scorer averages class F1 over the union of reference and predicted
labels, including \texttt{none} or \texttt{legitimate} when present,
with zero for undefined class scores. It is therefore not a macro
average restricted to the eight typologies. Published scores use
the stored, postprocessed labels. The released live parser does
not fully reproduce every historical free-text fallback, so exact
table reproduction uses these archived labels.

\begin{table*}[t]
\centering
\small
\setlength{\tabcolsep}{4pt}
\begin{tabular}{llrrrlrl}
\toprule
Model & M & $F1_{\rm HT}$ & $\mathrm{SD}_{\rm run}$ & $\mathrm{SE}_{\rm samp}$ & 95\% interval & Lift & 95\% interval \\
 & & (\%) & ($10^{-3}$ pp) & ($10^{-3}$ pp) & (\%) & & \\
\midrule
\multicolumn{8}{c}{\textit{HI-Small}} \\
GPT-OSS-20B & FS & 0.2360 & 2.34 & 2.02 & $[0.2254, 0.2493]$ & 0.959 & $[0.9165, 1.0136]$ \\
GPT-OSS-20B & ZS & 0.2457 & 0.76 & 0.21 & $[0.2448, 0.2490]$ & 0.999 & $[0.9950, 1.0117]$ \\
Qwen3.5-27B & FS & 0.2404 & 0.90 & 4.52 & $[0.2191, 0.2667]$ & 0.978 & $[0.8915, 1.0851]$ \\
Qwen3.5-27B & ZS & 0.2302 & 0.43 & 1.75 & $[0.2224, 0.2406]$ & 0.936 & $[0.9041, 0.9778]$ \\
Nem-Nano-30B & FS & 0.2454 & 1.21 & 0.92 & $[0.2407, 0.2525]$ & 0.998 & $[0.9786, 1.0261]$ \\
Nem-Nano-30B & ZS & 0.2460 & 0.10 & 0.07 & $[0.2458, 0.2485]$ & 1.000 & $[0.9990, 1.0097]$ \\
Qwen3.5-35B-A3B & FS & 0.2525 & 2.74 & 1.70 & $[0.2411, 0.2670]$ & 1.027 & $[0.9804, 1.0853]$ \\
Qwen3.5-35B-A3B & ZS & 0.2463 & 1.40 & 0.41 & $[0.2430, 0.2519]$ & 1.001 & $[0.9878, 1.0236]$ \\
GPT-OSS-120B & FS & 0.2494 & 5.64 & 0.90 & $[0.2445, 0.2567]$ & 1.014 & $[0.9941, 1.0434]$ \\
GPT-OSS-120B & ZS & 0.2457 & 0.00 & 0.16 & $[0.2454, 0.2482]$ & 0.999 & $[0.9976, 1.0086]$ \\
Nem-Super-120B & FS & 0.2474 & 2.14 & 1.59 & $[0.2359, 0.2620]$ & 1.006 & $[0.9591, 1.0649]$ \\
Nem-Super-120B & ZS & 0.2472 & 0.76 & 0.32 & $[0.2447, 0.2519]$ & 1.005 & $[0.9947, 1.0239]$ \\
Qwen3.5-397B & FS & 0.2507 & 0.79 & 0.94 & $[0.2464, 0.2573]$ & 1.019 & $[1.0015, 1.0458]$ \\
Qwen3.5-397B & ZS & 0.2478 & 0.35 & 0.36 & $[0.2465, 0.2514]$ & 1.007 & $[1.0019, 1.0217]$ \\
\midrule
\multicolumn{8}{c}{\textit{LI-Small}} \\
GPT-OSS-20B & FS & 0.0982 & 0.69 & 0.80 & $[0.0946, 0.1035]$ & 0.900 & $[0.8673, 0.9479]$ \\
GPT-OSS-20B & ZS & 0.1093 & 0.26 & 0.10 & $[0.1088, 0.1115]$ & 1.002 & $[0.9975, 1.0218]$ \\
Qwen3.5-27B & FS & 0.0965 & 0.45 & 1.71 & $[0.0886, 0.1068]$ & 0.884 & $[0.8124, 0.9790]$ \\
Qwen3.5-27B & ZS & 0.0878 & 0.11 & 0.82 & $[0.0846, 0.0926]$ & 0.805 & $[0.7755, 0.8486]$ \\
Nem-Nano-30B & FS & 0.1084 & 0.54 & 0.30 & $[0.1071, 0.1115]$ & 0.993 & $[0.9821, 1.0210]$ \\
Nem-Nano-30B & ZS & 0.1091 & 0.17 & 0.06 & $[0.1089, 0.1111]$ & 1.000 & $[0.9981, 1.0173]$ \\
Qwen3.5-35B-A3B & FS & 0.1067 & 1.44 & 0.74 & $[0.1017, 0.1137]$ & 0.978 & $[0.9324, 1.0419]$ \\
Qwen3.5-35B-A3B & ZS & 0.1088 & 0.59 & 0.30 & $[0.1069, 0.1126]$ & 0.997 & $[0.9799, 1.0314]$ \\
GPT-OSS-120B & FS & 0.1083 & 5.40 & 0.40 & $[0.1065, 0.1120]$ & 0.993 & $[0.9761, 1.0255]$ \\
GPT-OSS-120B & ZS & 0.1092 & 0.13 & 0.07 & $[0.1091, 0.1111]$ & 1.001 & $[1.0000, 1.0179]$ \\
Nem-Super-120B & FS & 0.1089 & 1.05 & 0.58 & $[0.1051, 0.1147]$ & 0.998 & $[0.9636, 1.0508]$ \\
Nem-Super-120B & ZS & 0.1096 & 0.57 & 0.21 & $[0.1082, 0.1128]$ & 1.004 & $[0.9917, 1.0337]$ \\
Qwen3.5-397B & FS & 0.1105 & 0.27 & 0.35 & $[0.1092, 0.1135]$ & 1.012 & $[1.0008, 1.0400]$ \\
Qwen3.5-397B & ZS & 0.1100 & 0.21 & 0.26 & $[0.1092, 0.1125]$ & 1.008 & $[1.0010, 1.0310]$ \\
\bottomrule
\end{tabular}
\caption{Sampling uncertainty for the 28 five-run mean LLM results. F1 and its interval endpoints are percentages; run SD and linearized sampling SE are in $10^{-3}$ percentage points. Lift is precision divided by full-test illicit prevalence. Intervals are conservative, cellwise 95\% confidence envelopes obtained from simultaneous hypergeometric bounds for 15 run--stratum counts (\Cref{app:estimation}). They condition on the frozen design, realized allocation, and five fixed prediction vectors; they do not include uncertainty over future inference runs. Coverage is not simultaneous across the 28 cells. Endpoints are rounded outward.}
\label{tab:llm_sampling_uncertainty}
\end{table*}

\subsection{Full Results and Between-Run Variation}
\label{app:ht_per_model}
\label{app:seed_std}

Table~\ref{tab:llm_ht_full} gives the full HI- and LI-Small results
behind the main HI-Small summary (Table~\ref{tab:llm_results}).
It adds HT precision, recorded-token ratio, and compact-set F1 run SD.
All means use stochastic runs stored under identifiers
$[42,123,456,789,1011]$. The original runner did not pass these
identifiers as seeds to the model API. They identify repeated
runs, not controlled random streams; the current release runner
does pass request seeds.

Weighting changes the point-estimate ordering: on \hiSmall,
Qwen3.5-397B \fs has the largest compact F1 ($50.6\%$), while
Qwen3.5-35B-A3B \fs has the largest HT estimate ($0.2525\%$).
Spearman correlations between compact and HT orderings are
$0.64$ on \hiSmall and $0.94$ on \liSmall; these descriptive
orderings do not establish population model ranks.
At full-test prevalence, precision is much smaller than recall,
so $F1=2PR/(P+R)\approx2P$. We therefore show F1 and precision
lift in the main table, and retain precision here for inspection.

\begin{table*}[t]
\centering
\small
\setlength{\tabcolsep}{5pt}
\begin{tabular}{llrrlrl}
\toprule
Split & Model & $\Delta F1_{\rm HT}$ & $\mathrm{SE}_{\rm paired}$ & 95\% interval & $\Delta$Lift & 95\% interval \\
 & & (pp) & ($10^{-3}$ pp) & (pp) & & \\
\midrule
HI-Small & GPT-OSS-20B & -0.0098 & 2.01 & $[-0.0244, 0.0056]$ & -0.039 & $[-0.100, 0.023]$ \\
HI-Small & Qwen3.5-27B & 0.0101 & 4.14 & $[-0.0235, 0.0468]$ & 0.042 & $[-0.095, 0.192]$ \\
HI-Small & Nem-Nano-30B & -0.0006 & 0.92 & $[-0.0082, 0.0073]$ & -0.002 & $[-0.034, 0.030]$ \\
HI-Small & Qwen3.5-35B-A3B & 0.0063 & 1.61 & $[-0.0118, 0.0252]$ & 0.026 & $[-0.048, 0.103]$ \\
HI-Small & GPT-OSS-120B & 0.0037 & 0.90 & $[-0.0041, 0.0119]$ & 0.015 & $[-0.017, 0.049]$ \\
HI-Small & Nem-Super-120B & 0.0002 & 1.56 & $[-0.0171, 0.0185]$ & 0.001 & $[-0.070, 0.076]$ \\
HI-Small & Qwen3.5-397B & 0.0029 & 0.80 & $[-0.0055, 0.0114]$ & 0.012 & $[-0.023, 0.047]$ \\
LI-Small & GPT-OSS-20B & -0.0111 & 0.79 & $[-0.0173, -0.0049]$ & -0.102 & $[-0.159, -0.045]$ \\
LI-Small & Qwen3.5-27B & 0.0086 & 1.44 & $[-0.0049, 0.0232]$ & 0.079 & $[-0.044, 0.213]$ \\
LI-Small & Nem-Nano-30B & -0.0007 & 0.28 & $[-0.0041, 0.0028]$ & -0.006 & $[-0.038, 0.026]$ \\
LI-Small & Qwen3.5-35B-A3B & -0.0022 & 0.68 & $[-0.0115, 0.0074]$ & -0.020 & $[-0.105, 0.068]$ \\
LI-Small & GPT-OSS-120B & -0.0009 & 0.38 & $[-0.0049, 0.0031]$ & -0.008 & $[-0.045, 0.029]$ \\
LI-Small & Nem-Super-120B & -0.0007 & 0.56 & $[-0.0082, 0.0070]$ & -0.006 & $[-0.075, 0.064]$ \\
LI-Small & Qwen3.5-397B & 0.0005 & 0.25 & $[-0.0036, 0.0046]$ & 0.004 & $[-0.033, 0.042]$ \\
\bottomrule
\end{tabular}
\caption{Few-shot minus zero-shot differences in five-run mean metrics. F1 differences and endpoints are percentage points; paired linearized SE is in $10^{-3}$ percentage points. The SE uses same-edge covariance across both prompts and all runs. Conservative 95\% envelopes use simultaneous bounds on 30 run--stratum counts and do not exploit pairing. Coverage is per comparison, not simultaneous across this table. Endpoints are rounded outward.}
\label{tab:llm_sampling_contrasts}
\end{table*}

\subsection{Sampling Uncertainty}
\label{app:llm_sampling}

Table~\ref{tab:llm_sampling_uncertainty} separates between-run
SD from design-based sampling SE and conservative $95\%$
intervals for the mean of five fixed-run metrics, using cached
predictions. Neither run SD nor sampling SE is uniformly larger;
the intervals remain nonzero
when observed within-stratum prediction variance is zero.

The largest upper lift endpoint is $1.0853$ on \hiSmall and
$1.0508$ on \liSmall (rounded outward). Some cellwise intervals
lie just above one: for example, Qwen3.5-397B few-shot has lift
$1.0188$ with interval $[1.0015,1.0458]$ on \hiSmall.
Near-baseline performance therefore does not imply equality with
predict-all. These are cellwise statements; we do not make a
simultaneous claim about which models outperform the reference.

Table~\ref{tab:llm_sampling_contrasts} reports FS--ZS differences
with paired sampling SEs and conservative confidence intervals.
All seven \hiSmall intervals include zero. On \liSmall, the
GPT-OSS-20B interval is negative: $\Delta F1=-0.0111$ pp,
with interval $[-0.0173,-0.0049]$ pp; the other six include zero.
These intervals do not establish equivalence between prompts,
and their coverage is per comparison.

\subsection{Few-Shot versus Zero-Shot}
\label{app:fewshot}

The per-prompt results in Table~\ref{tab:llm_ht_full} show that
few-shot demonstrations improve typology F1 for all seven models
on \hiSmall and six on \liSmall. The largest gains are
$10.7$ pp on \hiSmall and $8.7$ pp on \liSmall, both for
Qwen3.5-397B. Compact-set detection F1 changes
range from $-5.0$ to $+0.5$ pp on \hiSmall and $-4.5$ to $+0.5$ pp
on \liSmall. Recorded-token ratios do not increase; the largest
decreases are Qwen3.5-27B's $26.6$ pp on \hiSmall and $11.2$ pp
on \liSmall.
These are descriptive differences on the compact set.
Table~\ref{tab:llm_sampling_contrasts} instead reports HT-weighted
FS--ZS differences with sampling uncertainty.

\FloatBarrier

\section{Error Analysis: ML vs LLM}
\label{app:errors_group}

\subsection{ML Scores by Correctness Group}
\label{app:error_dist}

Table~\ref{tab:error_dist} extends the main decomposition to all
seven LLMs (few-shot, run~$42$). LLM-only-correct edges are mostly
illicit, while ML-only-correct edges are mostly benign. Both groups
have low mean ML scores; their ground-truth composition does not
provide an observable routing rule.

\begin{table}[!htbp]
\centering
\small
\setlength{\tabcolsep}{2pt}
\begin{tabular}{llrrcc}
\toprule
\textbf{LLM} & \textbf{Group} & \textbf{n} & \textbf{ill.} & \textbf{$\bar{s}$} & $\sigma_s$ \\
\midrule
\multicolumn{6}{c}{\textit{HI-Small}} \\
\midrule
GPT-OSS-20B & Both correct & 1,142 & 566 & .53 & .47 \\
 & ML only correct & 1,944 & 145 & .11 & .29 \\
 & LLM only correct & 407 & 366 & .17 & .28 \\
 & Both wrong & 260 & 174 & .31 & .41 \\
\midrule
Nem-Nano & Both correct & 802 & 700 & .86 & .30 \\
 & ML only correct & 2,284 & 11 & .06 & .18 \\
 & LLM only correct & 509 & 495 & .09 & .19 \\
 & Both wrong & 158 & 45 & .64 & .40 \\
\midrule
GPT-OSS-120B & Both correct & 963 & 652 & .69 & .44 \\
 & ML only correct & 2,123 & 59 & .08 & .22 \\
 & LLM only correct & 433 & 410 & .13 & .23 \\
 & Both wrong & 234 & 130 & .40 & .44 \\
\midrule
Nem-Super & Both correct & 1,023 & 596 & .60 & .46 \\
 & ML only correct & 2,063 & 115 & .10 & .27 \\
 & LLM only correct & 474 & 444 & .12 & .24 \\
 & Both wrong & 193 & 96 & .47 & .43 \\
\midrule
Qwen3.5-397B & Both correct & 818 & 710 & .87 & .29 \\
 & ML only correct & 2,268 & 1 & .05 & .16 \\
 & LLM only correct & 541 & 522 & .10 & .20 \\
 & Both wrong & 126 & 18 & .76 & .31 \\
\midrule
Qwen3.5-35B-A3B & Both correct & 1,053 & 622 & .60 & .47 \\
 & ML only correct & 2,033 & 89 & .09 & .25 \\
 & LLM only correct & 470 & 440 & .12 & .24 \\
 & Both wrong & 197 & 100 & .45 & .44 \\
\midrule
Qwen3.5-27B & Both correct & 1,359 & 354 & .29 & .43 \\
 & ML only correct & 1,727 & 357 & .25 & .40 \\
 & LLM only correct & 380 & 336 & .17 & .29 \\
 & Both wrong & 287 & 204 & .29 & .39 \\
\bottomrule
\end{tabular}
\caption{Detection-error groups on \hiSmall (few-shot, run~$42$).
$n$ and ill.\ are unweighted total and illicit counts;
$\bar{s}$ and $\sigma_s$ are the mean and std of the two-booster ML
score (threshold $0.80$). The main task results average five runs.}
\label{tab:error_dist}
\end{table}

\subsection{Typology Classification after Correct Detection}
\label{app:error_transitions}

\begin{figure*}[!t]
\centering
\includegraphics[width=\textwidth]{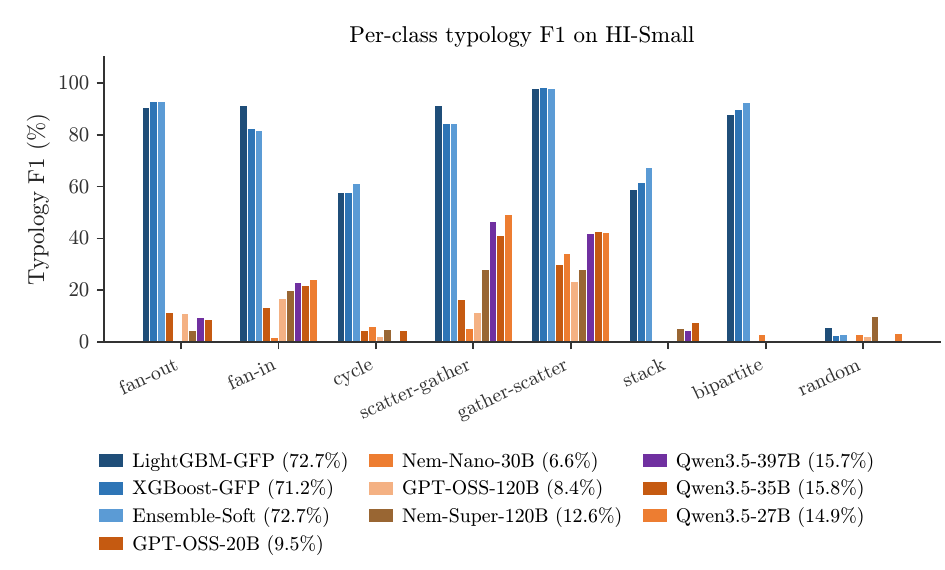}
\caption{Per-class typology $F1$ on \hiSmall (few-shot, run~$42$;
ML threshold $0.80$), conditional on correct illicit detection.
The ensemble's macro-$F1$ is $72.7\%$ versus $6.6$--$15.8\%$ for
LLMs. Unlike the TF1 in \Cref{tab:llm_results}, this comparison
uses each model's true positives. Class supports appear in
\Cref{tab:typology_prior}.}
\label{fig:typology_f1}
\end{figure*}

Figure~\ref{fig:typology_f1} compares typology classification among
correctly detected illicit edges. This differs from detection
recall by true typology in \Cref{tab:per_typology_recall}.

\subsection{Over-Prediction Ratio and Per-Typology Recall}
\label{app:over_predict_stats}

The over-prediction ratio is the predicted-illicit count divided by
the true-illicit count. For nonzero TP, it also equals recall divided
by precision.
For run~$42$, it spans $1.71$--$3.00$ on \hiSmall and
$2.11$--$3.00$ on \liSmall across models and prompting variants.
Under zero-shot prompting, the six models other than Qwen3.5-27B
span $2.94$--$3.00$ and $2.91$--$3.00$, respectively;
$3$ is the predict-all value for the $1{:}2$ subset.
Qwen3.5-27B's context-budget rejections suppress its flagged count.
These are run-$42$ diagnostics, distinct from the five-run means
in \Cref{tab:llm_ht_full}.

\Cref{tab:per_typology_recall} separates detection recall by true
typology to identify which patterns the models miss.

\begin{table}[!htbp]
\centering
\small
\setlength{\tabcolsep}{4pt}
\begin{tabular}{lrrr}
\toprule
\textbf{Typology} & \textbf{n} & \textbf{ML Ens.} & \textbf{LLM} \\
\midrule
fan-in           &  90 & \textbf{58.9} & 98.9 \\
bipartite        &  43 & \textbf{60.5} & 97.7 \\
random           &  66 & 84.8 & 97.0 \\
fan-out          &  85 & 94.1 & 88.2 \\
stack            &  32 & 96.9 & 100.0 \\
cycle            &  74 & 100.0 & 98.6 \\
scatter-gather   & 142 & 95.8 & 99.3 \\
gather-scatter   & 259 & 98.5 & 83.4 \\
\bottomrule
\end{tabular}
\caption{Per-typology recall (\%) on \hiSmall illicit edges with
programmatic labels ($n{=}791$ total). ML Ens.\ = soft-vote of
LightGBM and XGBoost at $\tau{=}0.80$; LLM = GPT-OSS-120B FS
(run~$42$). Bold = ML's weakest typologies, where LLMs
recover the missed illicit edges.}
\label{tab:per_typology_recall}
\end{table}

\subsection{Failure Taxonomy}
\label{app:failure_taxonomy}

Table~\ref{tab:failure_taxonomy} reports the per-model breakdown
referenced in~\Cref{sec:llm:traces}.

\begin{table}[!htbp]
\centering
\small
\setlength{\tabcolsep}{3pt}
\begin{tabular}{lrrrrr}
\toprule
\textbf{Model} & \textbf{Cor.} & \textbf{Over} & \textbf{Und.} & \textbf{Typo} & \textbf{NR} \\
\midrule
GPT-OSS-20B       & 17.5 & 50.2 & 1.8 & 13.5 & 17.0 \\
GPT-OSS-120B      & 15.5 & 57.8 & 0.2 & 16.2 & 10.4 \\
Nem-Nano-30B      & 18.9 & 63.6 & 0.6 & 14.9 & 2.0 \\
Nem-Super-120B    & 24.9 & 54.5 & 4.7 & 13.9 & 2.0 \\
Qwen3.5-35B-A3B   & 28.4 & 54.4 & 4.6 & 11.7 & 0.9 \\
Qwen3.5-397B      & 22.3 & 63.3 & 0.1 & 13.4 & 0.9 \\
Qwen3.5-27B       & 16.0 & 38.7 & 0.3 & 6.3  & 38.7 \\
\bottomrule
\end{tabular}
\caption{Failure taxonomy on \hiSmall (\fs), in percent of cases.
\textit{Cor.}=correct binary verdict and, where a reference exists,
correct typology; \textit{Over}=over-prediction (FP),
\textit{Und.}=under-prediction (FN), \textit{Typo}=wrong typology given
correct detection, \textit{NR}=no response, meaning the request returned
no output to score rather than output we failed to parse. For Qwen3.5-27B
these are API-layer rejections from a context-budget mismatch, not a
reasoning outcome (\Cref{app:examples}). All columns are unweighted percentages of the $3{,}753$ run-$42$
cases. NR counts requests with no returned response or reasoning
text; an empty final response with a nonempty reasoning trace
is not counted as NR.}
\label{tab:failure_taxonomy}
\end{table}

\subsection{Typology Concentrations by Model and Prompting}
\label{app:typology_freq}

The main heatmap (\Cref{fig:anchor_heatmap}) includes legitimate
and missing labels among all ground-truth-benign cases. Conditioning
instead on predicted illicit typologies, the dominant label's share
ranges from $35\%$ to $91\%$ on \hiSmall and from $42\%$ to $91\%$ on \liSmall
across the seven models under \fs and \zs, aggregated over five runs.
For GPT-OSS-120B on \hiSmall, fan-out accounts for $52\%$ under \fs and
gather-scatter for $70\%$ under \zs. Both Nemotron models favor
gather-scatter under \zs. Per-model shares and labels are included
in \path{results/analysis/anchor_chi_square.csv}; the ranges here
exclude that file's verification-prompt cells.

\FloatBarrier

\section{Doubt Triage: Methods and Results}
\label{app:dt_stats}

\subsection{Combining ML and LLM Predictions: A Diagnostic Reference}
\label{sec:hybrid}

Doubt Triage (DT) combines the two-booster ML reference and LLM
predictions using HT-Coreset strata.
These strata incorporate ground-truth labels, so DT is an
optimistic diagnostic reference rather than a deployable policy or
a mathematical upper bound over hybrid rules.

\subsubsection{Routing Rule}
\label{sec:hybrid:method}
\label{app:dt_algorithm}

Algorithm~\ref{alg:dt} specifies the routing rule. DT consults the
LLM only on ML-negative edges in the weight-one stratum
($w_i=1$: all illicit edges and retained hard-negative benign
edges) and uses ML elsewhere. Detection metrics use HT-weighted
counts. For DT and the
ensemble, these equal the unweighted metrics because neither flags
a sampled benign edge.
To measure the contribution of the routing rule, we also assign an
illicit label to every edge on this same DT route without querying an
LLM.

\begin{algorithm}[!ht]
\caption{Doubt Triage (DT)}
\label{alg:dt}
\begin{algorithmic}[1]
\REQUIRE Edge $e_i$, HT weight $w_i$, ML score $s(e_i)$, threshold
         $\tau$ ($0.80$ on \hiSmall; $0.48$ on \liSmall), ML ensemble
         $f_{\text{ML}}$, LLM judge $f_{\text{LLM}}$
\ENSURE  Predicted label $\hat{y}_i$, source $\in\{\text{ML}, \text{LLM}\}$
\IF{$w_i > 1$}
  \STATE \textbf{return} $(f_{\text{ML}}(e_i),\ \text{ML})$ \COMMENT{sampled stratum (easy benign)}
\ELSIF{$s(e_i) \geq \tau$}
  \STATE \textbf{return} $(f_{\text{ML}}(e_i),\ \text{ML})$ \COMMENT{census, ML-illicit}
\ELSE
  \STATE \textbf{return} $(f_{\text{LLM}}(e_i),\ \text{LLM})$ \COMMENT{census, ML-legit: defer}
\ENDIF
\end{algorithmic}
\end{algorithm}

ML-contributed true positives take the ML typology prediction;
LLM-recovered true positives take the LLM prediction.
\Cref{app:naive_hybrids} evaluates routing
without access to those strata.

\begin{table}[!t]
\centering
\small
\setlength{\tabcolsep}{2.5pt}
\begin{tabular}{lccccc}
\toprule
\textbf{Method} & \textbf{F1} & \textbf{P} & \textbf{R} & \textbf{TF1} & \textbf{\%C} \\
\midrule
\multicolumn{6}{c}{\textit{HI-Small (ML threshold = 0.80)}} \\
\midrule
ML Ensemble & 68.1 & 84.8 & 56.8 & 72.7 & 0 \\
Score-gate (14-cell mean) & 65.3 & 73.8 & 58.5 & 71.2 & 5 \\
Same-route always illicit & 87.3 & 77.5 & 100.0 & --- & 0 \\
\midrule
DT + GPT-OSS-20B & 87.3 & 77.6 & 99.7 & 57.0 & 21 \\
DT + Qwen3.5-27B & 82.0 & 75.9 & 89.1 & \textbf{59.7} & 21 \\
DT + Nem-Nano-30B & 87.3 & 77.5 & 100.0 & 58.6 & 21 \\
DT + Qwen3.5-35B-A3B & 86.8 & 77.4 & 98.8 & 56.8 & 21 \\
DT + GPT-OSS-120B & 87.2 & 77.5 & 99.8 & 56.4 & 21 \\
DT + Nem-Super-120B & 87.3 & 77.7 & 99.6 & 56.2 & 21 \\
DT + Qwen3.5-397B & \textbf{87.3} & 77.5 & \textbf{100.0} & 56.0 & 21 \\
\midrule
\multicolumn{6}{c}{\textit{LI-Small (ML threshold = 0.48)}} \\
\midrule
ML Ensemble & 28.6 & 60.2 & 18.8 & 66.2 & 0 \\
Score-gate (14-cell mean) & 34.2 & 50.8 & 25.8 & 57.1 & 7 \\
Same-route always illicit & 84.9 & 73.8 & 100.0 & --- & 0 \\
\midrule
DT + GPT-OSS-20B & 84.9 & 73.8 & 99.8 & 39.7 & 35 \\
DT + Qwen3.5-27B & 71.7 & 68.4 & 75.4 & 44.0 & 35 \\
DT + Nem-Nano-30B & 84.8 & 73.7 & 99.8 & \textbf{44.1} & 35 \\
DT + Qwen3.5-35B-A3B & 83.5 & 73.3 & 97.0 & 43.2 & 35 \\
DT + GPT-OSS-120B & \textbf{84.9} & 73.8 & \textbf{100.0} & 38.3 & 35 \\
DT + Nem-Super-120B & 84.6 & 73.8 & 99.0 & 40.7 & 35 \\
DT + Qwen3.5-397B & 84.9 & 73.8 & 99.9 & 38.7 & 35 \\
\bottomrule
\end{tabular}
\caption{DT results on \bench{} ($1{:}2$ subset ratio). All metrics are percentages.
DT rows use \zs and report means over $5$ runs
($F1$ std $<0.3$ pp; TF1 std up to $1.0$ pp on \liSmall
GPT-OSS-20B). Detection $F1$, P, and R use HT-weighted counts;
TF1 is typology macro-$F1$ on correctly detected illicit edges
with a valid typology output. \%C is the percentage of compact-set
edges whose rule uses an LLM prediction. The always-illicit control
uses DT's same eligible route ($21\%$ of \hiSmall edges and $35\%$
of \liSmall edges) but requires
no LLM calls
and supplies no typology on routed edges, so its TF1 is unreported.
DT uses strata constructed from ground-truth labels.
\emph{Score-gate} rows are means over $14$ separately
grid-searched model--prompting cells ($7$ models, FS and ZS,
run~$42$) for the score-only recall gate, using DT's typology
fusion. Their per-cell $F1$ ranges are $64.7$--$66.4\%$ on
\hiSmall and $33.4$--$37.2\%$ on \liSmall
(\Cref{app:naive_hybrids}).}
\label{tab:dt}
\end{table}

\subsubsection{Results}
\label{sec:hybrid:results}
\label{app:significance}

Table~\ref{tab:dt} reports per-model zero-shot results and two
routing controls. Across both prompting variants, DT exceeds the
same-route always-illicit control in $14/70$ cells per split; the
rest equal or fall below it. DT exceeds ML in all $140$ tested
(model, dataset, prompting, run) cells, which reuse the same edges.
This comparison does not isolate LLM value beyond the label-derived
route or establish generalization across datasets. Per-cell,
pooled, and cross-model tests are in \path{results/stats/}; they
do not establish equivalence across models or prompting variants.

The score-only rows use the evaluation-selected gates specified in
\Cref{app:naive_hybrids}; their run scope differs from the DT rows.

On \hiSmall, TP-conditional typology macro-$F1$ is $72.7\%$ for
the ensemble and $57.2\%$ for DT averaged over the seven zero-shot
models. Their true-positive sets differ: detections recovered by
DT receive the LLM's weaker typology predictions.

\paragraph{Consultation cost.}
Table~\ref{tab:per_pass_tokens} compares a compact-set pass with
the estimated token cost of DT's consulted edges.

\begin{table}[!ht]
\centering
\small
\setlength{\tabcolsep}{4pt}
\begin{tabular}{@{}lrrrr@{}}
\toprule
\textbf{Model} & \multicolumn{2}{c}{\textbf{Compact pass}} & \multicolumn{2}{c}{\textbf{DT estimate}} \\
\cmidrule(lr){2-3}\cmidrule(lr){4-5}
& \textbf{HI} & \textbf{LI} & \textbf{HI} & \textbf{LI} \\
\midrule
GPT-OSS-20B          & $107$ & $\phantom{0}67$ & $22$ & $23$ \\
GPT-OSS-120B         & $106$ & $\phantom{0}66$ & $22$ & $23$ \\
Qwen3.5-27B          & $\phantom{0}98$ & $\phantom{0}51$ & $20$ & $18$ \\
Nem-Nano-30B         & $154$ & $\phantom{0}95$ & $32$ & $33$ \\
Nem-Super-120B       & $156$ & $\phantom{0}96$ & $32$ & $33$ \\
Qwen3.5-35B-A3B      & $165$ & $102$ & $34$ & $35$ \\
Qwen3.5-397B-A17B    & $159$ & $\phantom{0}98$ & $33$ & $34$ \\
\midrule
Range & $98$--$165$ & $51$--$102$ & $20$--$34$ & $18$--$35$ \\
\bottomrule
\end{tabular}
\caption{Model-token counts per compact-set pass (millions, run~42, \zs).
A compact-pass entry is (mean input $+$ mean output tokens) $\times$
the coreset size. DT estimates scale the unrounded full-pass cost
by $776/3{,}753$ on \hiSmall and $789/2{,}268$ on \liSmall
($21\%$ and $35\%$ consultation, respectively), assuming consulted edges have the
same mean token cost as the full set. This gives estimated savings
of $79\%$ on \hiSmall and $65\%$ on \liSmall; the relative spread
across models is unchanged.}
\label{tab:per_pass_tokens}
\end{table}

\subsection{Score-Only and Na\"ive Hybrid Rules}
\label{app:naive_hybrids}

An OR-rule predicts illicit whenever either model flags an edge.
At full-test prevalence, each additional LLM false positive in a
sampled benign stratum enters the estimated FP count at its HT
weight (about $470$ on \hiSmall). An AND-rule instead requires
both models to flag an edge and cannot recover an illicit edge
missed by ML. DT uses the evaluation stratum to restrict LLM
consultation, which requires information unavailable to these rules.

\paragraph{Tested score-only gate.}
We grid-search a recall-oriented deferral band over ML scores for
each model--prompt cell, using run~$42$ predictions and the same
typology fusion as DT. Table~\ref{tab:dt} reports means over $14$ cells per split;
the best cells reach $66.4\%$ and $37.2\%$ F1.
Thresholds are selected on the evaluation predictions, so these
are the best tested gates on that set, not held-out policy estimates. The original recall
grid runs from $0.001$ to $0.496$ in steps of $0.005$. On \hiSmall,
it excludes no deferral and all narrower bands between $0.496$
and the $0.80$ ML threshold, so its below-ML mean does not rule
out better score-only gates. Table~\ref{tab:dt}
distinguishes these single-run means from the five-run DT rows.

For the selected recall gates, HT-weighted and unweighted F1 agree:
the deferral bands exclude the sampled benign strata. Widening
the bands to include those strata can reduce weighted F1 below
$1\%$ through the added FP estimates. The selected gates still
recover few of the illicit edges missed by ML, whose scores overlap
with benign cases (\Cref{sec:llm:asymmetry}). The difference from
DT measures the advantage of using label-informed strata in this
benchmark; it does not establish a bound on other routing methods.

\FloatBarrier

\section{Reasoning-Trace Analysis}
\label{app:traces_group}

\begin{table*}[!ht]
\centering
\small
\setlength{\tabcolsep}{5pt}
\begin{tabular}{lrrrrrrrr}
\toprule
& \multicolumn{4}{c}{\textbf{Pass rate (\%)}} & \multicolumn{4}{c}{\textbf{Agreement ($\kappa$)}} \\
\cmidrule(lr){2-5}\cmidrule(lr){6-9}
\textbf{Step} & Regex & DeepSeek & Opus & Gemini & Fleiss & Mean pair & Min pair & Max pair \\
\midrule
Parse    & 80.4 & 83.5 & 73.9 & 81.3 & 0.80 & 0.81 & 0.71 & 0.91 \\
Recall   & 82.8 & 80.7 & 76.6 & 81.8 & 0.90 & 0.90 & 0.81 & 0.97 \\
Match    & 75.7 & 80.1 & 70.3 & 73.7 & 0.64 & 0.64 & 0.51 & 0.77 \\
Conclude & 19.2 & 24.0 & 27.1 & 20.9 & 0.73 & 0.74 & 0.66 & 0.87 \\
\bottomrule
\end{tabular}
\caption{Original step-rubric results on the same $1{,}000$ \hiSmall
traces (few-shot, run~$42$), including empty traces. Rates pool all
seven models; agreement uses item-level labels from four annotators
and six pairs. The regex heuristic uses full traces and API judges
use the first $8{,}000$ characters.}
\label{tab:step_rubric}
\label{tab:multi_judge_iaa}
\end{table*}

The $1{,}000$-trace regex audit draws up to $35$ traces per model
from each of four outcome groups (correct-illicit, over-prediction,
under-prediction, typology-error), then fills remaining slots from
unsampled model--case pairs. Correct-benign cases enter only through
this top-up. The released slice contains $183$ correct-illicit,
$9$ correct-benign, $295$ over-prediction, $235$ under-prediction,
and $278$ typology-error records; these are sampling outcomes, not
population shares. Of the typology-error records, $112$ lack a
reference typology, so their label alone cannot verify a typed
mismatch.

\subsection{Automated Audit by Model and Annotator}
\label{app:step_heatmap}
\label{app:multi_judge}

\begin{figure}[!htbp]
\centering
\includegraphics[width=\columnwidth]{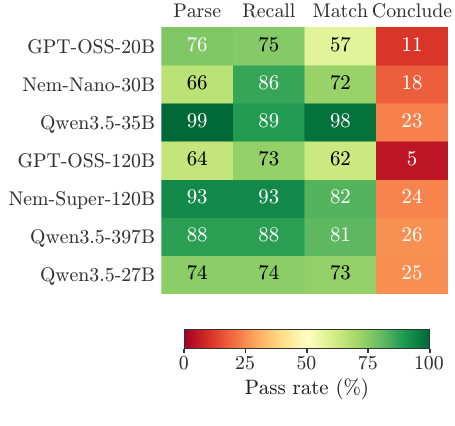}
\caption{Regex step-rubric pass rates by model on the $1{,}000$-trace
\hiSmall audit slice (\fs, run~$42$). Pooled annotator rates and
agreement appear in Table~\ref{tab:multi_judge_iaa}.}
\label{fig:step_heatmap}
\end{figure}

Figure~\ref{fig:step_heatmap} separates the pooled regex rates in
Table~\ref{tab:multi_judge_iaa} by model. The audit includes $148$
empty traces; the API judges' text window truncates $400$ traces.
The API judges are DeepSeek-V4-Pro, Claude Opus~4.7, and
\texttt{gemini-3-pro-preview}; the last differs from the Gemini~3.1
probe in \Cref{app:probe}. Fleiss' $\kappa$~\citep{fleiss1971agreement}
summarizes agreement across all four annotators; Cohen's $\kappa$~\citep{cohen1960agreement}
is computed for each of the six pairs. All four Fleiss' values
indicate at least substantial agreement~\citep{landis1977measurement}.

\subsection{Sensitivity to Scoring Rules and Text Windows}
\label{app:audit_sensitivity}

We recompute the frozen regex checks on full traces and their
first $8{,}000$ characters. All original full-trace labels reproduce
exactly. The pre-existing canonical typology-name mapping changes
no Recall decisions in either window. The extraction rules are
fixed independently of judge ratings.

The original Match rule passes every nonempty trace with a benign
verdict, even without two marker categories. Requiring those categories for all
verdicts removes this bypass. Separately, reference-defined answer
correctness is available for $456$ annotated illicit and $304$
benign records; the other $240$ illicit records have no reference
typology. Table~\ref{tab:audit_sensitivity_rules} reports both
changes and both text windows. Binary correctness remains
available for all records. Table~\ref{tab:audit_sensitivity}
conditions the revised lexical checks on the $470$ binary-correct
and $530$ binary-incorrect records. All denominators include empty
traces, which fail the lexical conjunction.

\begin{table}[!ht]
\centering\small
\setlength{\tabcolsep}{3pt}
\begin{tabular}{llrr}
\toprule
\textbf{Checks} & \textbf{References} & \textbf{Full} & \textbf{First 8k chars} \\
\midrule
Original regex & All & 538/1000 & 531/1000 \\
Original regex & Available & 421/760 & 417/760 \\
No bypass & All & 533/1000 & 525/1000 \\
No bypass & Available & 417/760 & 412/760 \\
\bottomrule\end{tabular}
\caption{Traces passing the specified lexical checks but failing the stated answer
checks. Rows marked All retain archived Conclude failures, including missing-reference cases;
Available-reference rows exclude all $240$ illicit cases without a
reference typology; benign cases remain evaluable. No-bypass rows require two
structural-marker categories even for benign verdicts. Original labels and scores
remain unchanged; each row applies a stated sensitivity rule to the same archived outputs.}
\label{tab:audit_sensitivity_rules}
\end{table}

The original $538$-record conjunction includes $303$ binary
errors ($237$ false positives and $66$ false negatives) and $235$
labeled typology errors, of which $97$ lack a reference typology.
After removing the benign-verdict bypass, $298$ binary-error
records still pass the lexical checks; this falls to $295$ in
the shorter window. These comparisons describe the sampled
records, not a population prevalence or a causal failure stage.

For correctness requiring both the binary verdict and, on illicit
cases, the typology, excluding the $240$ illicit records without a
reference typology leaves $760$ evaluable records. Of these, $417$
pass the revised lexical checks but fail answer correctness; with
the shorter window, the count is $412$
(Table~\ref{tab:audit_sensitivity_rules}).

The API judges receive the trace, prediction, and ground-truth
labels, but no transaction graph. They therefore do not
independently verify factual grounding in the graph. Their
Conclude scores differ from the frozen parser on $86$ records
for DeepSeek, $123$ for Opus, and $73$ for Gemini. These are
scoring-rule disagreements. The released analysis reports results
for all $1{,}000$ records and separately for the $852$ nonempty
traces. Per-model and outcome-conditioned counts,
compact per-item features, and a script reproducing all reports
without model calls accompany the release.

\appendixkeepheading{9\baselineskip}
\subsection{Human Ratings of the Step Rubric}
\label{app:human_rating}

\begin{table}[!htbp]
\centering
\small
\setlength{\tabcolsep}{3pt}
\begin{tabular}{lccccc}
\toprule
\textbf{Step} & \textbf{NLP} & \textbf{Jr.} & \textbf{Sr.} & \textbf{Mean} & \textbf{Fleiss' $\kappa$} \\
\midrule
Parse    & 100.0 & 100.0 & 96.4 & 98.8 & $-0.01$ \\
Recall   & 100.0 & 100.0 & 100.0 & 100.0 & undefined$^\dagger$ \\
Match    & 53.6 & 96.4 & 67.9 & 72.6 & 0.10 \\
Conclude & 21.4 & 21.4 & 21.4 & 21.4 & 1.00 \\
\bottomrule
\end{tabular}
\caption{Human pass rates (\%) and agreement on $28$ traces.
Parse has $27/28$ unanimous items but expected agreement of $0.976$,
yielding $\kappa=-0.01$. $^\dagger$Recall is uniformly positive.
Conclude agreement is imposed by the shared parser score.}
\label{tab:human_passrates}
\label{tab:human_iaa}
\end{table}

\begin{table*}[!t]
\centering
\small
\setlength{\tabcolsep}{3pt}
\begin{tabular}{lrrrrrrrrr}
\toprule
& \multicolumn{3}{c}{\textbf{Few-shot}} & \multicolumn{3}{c}{\textbf{Zero-shot}} & \multicolumn{3}{c}{\textbf{Few-shot by correctness}} \\
\cmidrule(lr){2-4}\cmidrule(lr){5-7}\cmidrule(lr){8-10}
\textbf{Model} & \textbf{Rate} & \textbf{Mean} & \textbf{Median} & \textbf{Rate} & \textbf{Mean} & \textbf{Median} & \textbf{Correct} & \textbf{Wrong} & \textbf{$\Delta$} \\
\midrule
GPT-OSS-20B     & 83.0 & 4{,}147 & 3{,}438 & 99.8 & 2{,}179 & 2{,}013 & 4{,}176 & 4{,}129 & $-47$ \\
Qwen3.5-27B     & 61.3 & 14{,}764 & 14{,}678 & 88.1 & 12{,}647 & 12{,}125 & 14{,}280 & 15{,}041 & $+761$ \\
Nem-Nano-30B    & 98.0 & 2{,}810 & 2{,}343 & 100.0 & 1{,}485 & 1{,}441 & 2{,}853 & 2{,}788 & $-66$ \\
Qwen3.5-35B-A3B & 99.1 & 22{,}208 & 19{,}204 & 100.0 & 15{,}566 & 14{,}585 & 23{,}980 & 21{,}003 & $-2{,}977$ \\
GPT-OSS-120B    & 89.6 & 1{,}998 & 1{,}899 & 99.8 & 1{,}147 & 1{,}117 & 1{,}965 & 2{,}015 & $+51$ \\
Nem-Super-120B  & 98.0 & 6{,}458 & 4{,}161 & 100.0 & 2{,}930 & 2{,}144 & 6{,}604 & 6{,}361 & $-243$ \\
Qwen3.5-397B    & 99.1 & 11{,}578 & 10{,}894 & 100.0 & 10{,}456 & 10{,}149 & 11{,}199 & 11{,}792 & $+593$ \\
\bottomrule
\end{tabular}
\caption{Trace statistics on \hiSmall (run~$42$). Rate is the
percentage of cases with nonempty \texttt{reasoning\_content};
all lengths are character counts of these traces. Correct and Wrong
give the mean lengths for correct and incorrect binary verdicts
under \fs; $\Delta$ reports wrong minus correct.
Displayed lengths and differences are rounded.}
\label{tab:trace_length}
\label{tab:trace_correctness}
\end{table*}

Three annotators---one NLP researcher and two AML practitioners
(junior and senior)---independently rated $28$ \hiSmall traces
(run~$42$) using the same rubric prompt as the API judges.
The slice contains one trace per model--outcome cell: seven models
and four outcomes (correct-illicit, over-prediction, typology-error,
under-prediction). For all raters, we replace Conclude with the
shared parser-derived score: one for a correct benign verdict, or
an illicit verdict with a present, matching typology reference.
Missing illicit references remain failures under this archived
convention. The released pack
contains the adopted rating sheets, including the senior rater's
calibration rescore; superseded sheets remain in the authors' archive.

Human Match agreement is low ($\kappa=0.10$). In particular, the
NLP rater marks non-suspicious verdicts on illicit edges as Match
failures, whereas the AML raters credit structural markers
independently of the verdict. Mean Match rates by outcome are
$81.0\%$, $57.1\%$, $81.0\%$, and $71.4\%$ in the order above.
Parse and Recall are uniformly positive except for Parse on
under-predictions ($95.2\%$). Under the shared parser rule,
six of the seven traces assigned to correct-illicit pass Conclude
($85.7\%$); all three error groups score zero.

These outcome-stratified ratings assess rubric application on a
different sample from the $1{,}000$-record automated audit. They do
not estimate population rates or paired human--judge disagreement.

\subsection{Trace-Length Statistics}
\label{app:trace_stats}

Table~\ref{tab:trace_length} combines trace availability and length
under both prompting methods with few-shot lengths conditioned on
binary correctness; \Cref{sec:llm:traces} interprets the differences.

\subsection{Prediction and Deployment Error Examples}
\label{app:examples}

These three \hiSmall cases (\fs, run~$42$) illustrate prediction
and deployment errors. The interpretations below are qualitative;
the quantitative rubric is defined in \Cref{sec:llm:failure}.
Quoted excerpts are from \texttt{reasoning\_content} or
\texttt{raw\_response}, with [\ldots] marking omissions.

\paragraph{Case 1: Different typologies from the same graph
(\texttt{amlc\_01031}, GT: \emph{cycle}).}
Both GPT-OSS-120B and Qwen3.5-397B label this $950$-transaction
subgraph suspicious. GPT-OSS-120B selects \emph{cycle}, describing
accounts that ``send funds back or forward in tightly-linked loops''
and ``back-and-forth movements across multiple banks''.
Qwen3.5-397B selects \emph{scatter-gather}, describing
\texttt{acct\_100428A08} as a ``high-volume distribution hub with
163 outgoing transactions'' and noting ``repetitive transaction
amounts''.
Both traces identify the hub and repeated transfers but emphasize
different structures when assigning a typology.

\paragraph{Case 2: A false positive using AML vocabulary
(\texttt{amlc\_00090}, GT: \emph{benign}).}
GPT-OSS-120B labels this $526$-transaction subgraph suspicious
(\emph{stack}, confidence $0.8$). Its explanation describes a
``dominant hub'' with an ``extremely high out-degree'' and states:
\begin{quote}\small
``This fan-out / gather-scatter pattern, with structured repeated
amounts and tight transaction timing, matches th[e laundering profile].''
\end{quote}
The trace names graph features and laundering patterns while giving
an incorrect verdict. This illustrates the observed gap between
such textual markers and prediction correctness; it does not
establish which internal computation caused the error.

\paragraph{Case 3: Request rejection
(\texttt{amlc\_03159}, Qwen3.5-27B, GT: \emph{gather-scatter}).}
The vLLM server returns HTTP~400 before inference because the
requested input and output exceed the deployed context cap:
\mbox{$98{,}306 + 32{,}767 = 131{,}073 > 131{,}072$} tokens.
Of this model's $1{,}453$ empty-output cases, $1{,}443$ ($99.3\%$)
carry the same error. The model supports a native $256$K context
(Table~\ref{tab:models}), but we deployed it at $128$K because of memory
constraints under four-way tensor parallelism. This explains most
of its $38.7\%$ no-response rate in Table~\ref{tab:failure_taxonomy};
these rejected requests do not provide a reasoning trace to audit.

\FloatBarrier

\section{Prompt Experiments and Templates}
\label{app:prompts}

\subsection{Main Evaluation Prompts}
We reuse the preamble, task wording, and answer format of
\citet{pirmorad2025iclaml} and follow its $8{+}4$ demonstration
structure. We draw the twelve example graphs from our training
split and write or adapt their brief explanations; the resulting
block is approximately $30$K tokens. \zs omits the examples.
The executed random-typology demonstration describes sub-threshold
structuring rather than AMLworld's random-walk pattern; we retain
that prompt in all reported results. The following sections describe
the additional prompt variants.

\subsection{Verification Prompt}
\label{app:intervention}

Table~\ref{tab:intervention} compares the verification prompt
(\verif) with \fs for two models on both compact evaluation sets
(run~$42$).

\paragraph{\verif verification block.}
Insert the block below before the answer-format header and add
``must reference at least one structural marker if Suspicious''
to the Explanation field. The rest of \fs is unchanged.

\begin{promptbox}[breakable=false,left=3pt,right=3pt,top=2pt,bottom=2pt,title=\verif: Verification step]
\small\ttfamily
Verification step (perform before concluding):\\
1. If the subgraph shows a hub account with recurring
fixed-amount transfers to many recipients with regular
timestamps (e.g., monthly/biweekly), this pattern is consistent
with payroll/salary disbursement and is NOT suspicious. Do not
select fan-out, gather-scatter, or stack solely on degree and
volume features.\\
2. If the only observed signal is high transaction volume,
identify at least one additional structural marker (cycles,
identical amounts crossing reporting thresholds, or rapid
layering across multiple counterparties) before classifying as
suspicious.\\
3. If you cannot identify a specific structural marker
matching one of the 8 typologies, classify the transaction as
Not Suspicious.
\end{promptbox}

\begin{table}[H]
\centering
\small
\setlength{\tabcolsep}{2pt}
\begin{tabular}{llrrrrr}
\toprule
\textbf{Model} & \textbf{D} & \textbf{F1} & \textbf{P} & \textbf{R} & \textbf{Over} & \textbf{Und} \\
\midrule
GPT-OSS-120B   & HI & $-0.94$ & $+0.12$ & $-6.4$  & $-4.6$  & $+2.1$ \\
GPT-OSS-120B   & LI & $+0.79$ & $+1.41$ & $-4.1$  & $-6.6$  & $+1.4$ \\
Nem-Super-120B & HI & $-3.00$ & $+2.16$ & $-22.9$ & $-18.6$ & $+7.6$ \\
Nem-Super-120B & LI & $-2.23$ & $+2.75$ & $-25.1$ & $-21.3$ & $+8.4$ \\
\midrule
Pooled mean    & -- & $-1.35$ & $+1.61$ & $-14.6$ & $-12.8$ & $+4.9$ \\
\bottomrule
\end{tabular}
\caption{Verification prompt effect (\verif $-$ \fs), in percentage
points of unweighted compact-set metrics (run~$42$).
D=dataset (HI or LI); Over/Und use FP/FN counts divided by all
cases in the compact set.
Precision rises and recall falls in all four cells; $F1$ changes
have mixed signs.}
\label{tab:intervention}
\end{table}

\begin{table*}[!ht]
\centering
\small
\setlength{\tabcolsep}{4pt}
\begin{tabular}{llrrrrr}
\toprule
\textbf{Model} & \textbf{Variant} & \textbf{Det-F1} & \textbf{Det-P} & \textbf{Det-R} & \textbf{Typ-F1} & \textbf{Typ-Acc} \\
\midrule
\multirow{6}{*}{DeepSeek-V4-Pro}
  & ZS-Graph        & $89.3$ & $88.8$ & $89.8$ & $20.9$ & $26.3$ \\
  & ZS-Base        & $89.0$ & $88.7$ & $89.2$ & $17.3$ & $21.7$ \\
  & FS-Base       & $86.4$ & $90.1$ & $83.0$ & $26.5$ & $28.3$ \\
  & ZS-Abstain    & $51.2$ & $90.0$ & $35.8$ & $16.1$ & $21.7$ \\
  & FS-CoT-Elim   & $82.3$ & $88.8$ & $76.7$ & $21.3$ & $23.7$ \\
  & FS-TypFirst   & $90.2$ & $89.0$ & $91.5$ & $\mathbf{30.3}$ & $\mathbf{32.3}$ \\
\midrule
\multirow{6}{*}{Gemini 3.1 Pro}
  & ZS-Graph        & $81.5$ & $87.6$ & $76.1$ & $24.5$ & $26.8$ \\
  & ZS-Base        & $80.7$ & $87.4$ & $75.0$ & $23.5$ & $26.3$ \\
  & FS-Base       & $82.4$ & $90.5$ & $75.6$ & $25.5$ & $27.3$ \\
  & ZS-Abstain    & $37.6$ & $\mathbf{97.6}$ & $23.3$ & $11.9$ & $20.7$ \\
  & FS-CoT-Elim   & $70.5$ & $88.8$ & $58.5$ & $19.5$ & $24.2$ \\
  & FS-TypFirst   & $85.5$ & $87.5$ & $83.5$ & $27.3$ & $27.8$ \\
\midrule
\multirow{6}{*}{Claude Sonnet 4.6}
  & ZS-Graph        & $91.8$ & $88.8$ & $94.9$ & $12.4$ & $18.7$ \\
  & ZS-Base        & $91.5$ & $88.8$ & $94.3$ & $12.5$ & $18.2$ \\
  & FS-Base       & $90.8$ & $89.5$ & $92.0$ & $21.0$ & $26.3$ \\
  & ZS-Abstain    & $80.4$ & $92.6$ & $71.0$ & $18.0$ & $23.7$ \\
  & FS-CoT-Elim   & $66.9$ & $91.2$ & $52.8$ & $17.9$ & $21.2$ \\
  & FS-TypFirst   & $\mathbf{93.5}$ & $89.6$ & $\mathbf{97.7}$ & $21.1$ & $25.8$ \\
\bottomrule
\end{tabular}
\caption{Frontier API probe on $198$ cases: $22$ per class across
$8$ illicit typologies and benign ($8{:}1$ illicit-to-benign ratio).
All metrics are percentages. Typology metrics use this nine-class pool;
detection uses the parsed typology, not the stated binary verdict.
Missing typologies ($249$ of $3{,}564$ responses) count as benign in
all metrics. The $18$ scored cells have no API errors. Bold marks
column maxima across those cells. Instructions and exact template
stems appear in \Cref{app:probe_prompts}.}
\label{tab:probe-frontier-apis}
\end{table*}

\appendixkeepheading{8\baselineskip}

\subsection{Frontier API Probe: Typology Prediction}
\label{app:probe}

To examine typology prediction beyond the open-weight models in
\Cref{sec:llm}, we evaluate six prompt variants on DeepSeek-V4-Pro,
Gemini~3.1~Pro Preview, and Claude Sonnet~4.6. All request at most
$16{,}384$ response tokens. Gemini and Sonnet also request $8{,}192$
thinking tokens; DeepSeek sets no separate thinking budget.
The pool contains $198$ \hiSmall compact-set cases, $22$ per
class ($8$ illicit typologies and benign). The prompts differ from
the main \fs/\zs in instructions and graph inputs; the $8{:}1$
class mix and scoring also differ. Comparisons concern variants
within this probe.

\paragraph{Input and scoring protocol.}
All variants remove self-transfers and collapse repeats into a count
and mean amount. Only the three few-shot variants add account roles
(SOURCE, SINK, HUB, PASS) and computed structural features.
Detection follows the parsed typology
(Table~\ref{tab:probe-frontier-apis}).

\paragraph{Prompt comparisons.}
\emph{FS-TypFirst} has the highest Typ-$F1$ among the six tested
variants for each API. Relative to \emph{ZS-Graph}, \emph{ZS-Abstain}
reduces detection recall and reaches precision of at least $90\%$.

\paragraph{Typology errors remain frequent.}
The highest typology macro-$F1$ is $30.3\%$ (DeepSeek-V4-Pro
\emph{FS-TypFirst}). Sonnet~4.6 \emph{FS-TypFirst} reaches
$93.5\%$ detection $F1$ and $21.1\%$ typology $F1$.
The probe does not isolate the effects of pre-training, model size,
or thinking budget.

\appendixkeepheading{8\baselineskip}

\subsubsection{Prompt Specifications}
\label{app:probe_prompts}

System and user templates for all six variants are in
\path{prompts/frontier_probe/}.

\begin{itemize}[label={},leftmargin=0pt,itemsep=4pt,parsep=0pt]
\item \textbf{ZS-Graph}: No demonstrations; classify the subgraph
as suspicious or not suspicious.
\item \textbf{ZS-Base}: No demonstrations; use ZS-Graph's system
prompt and ask about the flagged transaction in its surrounding graph.
\item \textbf{FS-Base}: Use $8$ suspicious and $4$ non-suspicious
examples and a computed test-case features header.
\item \textbf{ZS-Abstain}: Give an initial
verdict and HIGH, MEDIUM, or LOW confidence. LOW forces a benign
verdict and \texttt{legitimate} pattern. Otherwise, an illicit verdict
requires at least three of four conditions: two specific features
matching one typology; evidence beyond hub structure; HIGH or MEDIUM
confidence in one typology; and ruling out plausible legitimate-business
explanations.
\item \textbf{FS-CoT-Elim}: Share FS-Base's
system prompt and demonstrations. Mark all eight typologies REJECT
or CONSIDER with structural reasons. Return benign if none survives;
otherwise select the one with the strongest evidence.
\item \textbf{FS-TypFirst}: Use $8$ suspicious
examples and no benign examples for an upstream-flagged case.
Return Explanation, Observed Pattern, then Conclusion; derive the
verdict from the illicit or \texttt{legitimate} pattern.
\end{itemize}

\end{document}